\documentclass{article} 
\usepackage{iclr2027_conference,times}

\usepackage{amsmath,amsfonts,bm}

\def\eqref#1{equation~\ref{#1}}

\def\1{\bm{1}}

\DeclareMathAlphabet{\mathsfit}{\encodingdefault}{\sfdefault}{m}{sl}
\SetMathAlphabet{\mathsfit}{bold}{\encodingdefault}{\sfdefault}{bx}{n}

\usepackage{hyperref}
\usepackage{cleveref} 
\usepackage{url}

\usepackage{booktabs}
\usepackage{multirow}
\usepackage{amsmath}
\usepackage{amssymb}
\usepackage{mathtools}
\usepackage{graphicx}
\usepackage{xcolor}
\usepackage{algorithm}
\usepackage{algorithmic}
\usepackage{enumitem}
\usepackage{bbm}
\usepackage{comment}
\usepackage{pifont}
\usepackage[table]{xcolor}
\usepackage{soul}
\usepackage{wrapfig}
\usepackage{subcaption}
\usepackage{array}
\usepackage{caption}

\usepackage{tikz}
\usetikzlibrary{positioning,arrows.meta,fit}

\definecolor{otterbrown}{HTML}{7A5C3E}
\definecolor{otterdark}{HTML}{4F3A2A}
\definecolor{ottersand}{HTML}{E8D3B5}
\definecolor{ottercream}{HTML}{F7F1E8}
\definecolor{otterblue}{HTML}{6F9A9A}
\definecolor{otterlightblue}{HTML}{E8F1F0}
\definecolor{midgray}{HTML}{666666}

\definecolor{rowgray}{gray}{0.92}

\renewcommand{\paragraph}{\textbf}

\newcommand{\virg}[1]{``#1''}

\title{From What to Which: Decoding Modifier Grounding in Frozen MLLMs}

\author{%
  Barbara Toniella Corradini \\
  AI for Good (AIGO), Istituto Italiano di Tecnologia, Italy \\
  \texttt{barbara.corradini@iit.it} \\
  \AND 
  Caterina Gallegati \\
  University of Siena, Italy \\
  \texttt{c.gallegati@student.unisi.it} \\
  \AND 
  Ludovica Genovese \\
  AI for Good (AIGO), Istituto Italiano di Tecnologia, Italy \\
  University of Genoa, Italy \\
  \texttt{ludovica.genovese@iit.it}
  \AND
  Vittorio Murino \\
  AI for Good (AIGO), Istituto Italiano di Tecnologia, Italy \\
  University of Verona, Italy \\
  \texttt{vittorio.murino@iit.it} \\
}

\begin{document}
\iclrfinalcopy 
\maketitle

\begin{abstract}
As Multimodal Large Language Models (MLLMs) can describe increasingly complex visual scenes, token-level grounding becomes crucial. 
Yet, when an MLLM generates ``the yellow banana on the left'', established grounding approaches focus on \emph{what} is in the image (``banana''), overlooking tokens that help describe \emph{which} instance is meant (``yellow'', ``left'').
In this work, we ask whether frozen MLLM representations contain decodable grounding information about the referred instance across generated tokens, extending to modifiers such as attributes, spatial expressions, and relational/action terms.
To address this question, we introduce OTTER, a lightweight supervised probe over frozen MLLM representations that uses Optimal Transport (OT) to align generated tokens with visual regions and produce compact grounding maps.
Our results show that (i) instance-discriminative visual information can be decoded from modifier tokens, with the clearest evidence for spatial terms, but (ii) is not confined to them, as contextualized object nouns also carry referential information;
(iii) the recovered grounding remains informative under context perturbations, while selected regions remain relevant to generation;
and (iv) the learned OT-based grounding extends beyond the controlled setting to free generation and cross-dataset transfer.
\end{abstract}

\newcommand{\VM}[1]{\textcolor{blue}{\textbf{VM:} #1}}

\section{Introduction}
Recent advances in Multimodal Large Language Models (MLLMs) have made it increasingly important to understand how their generated language is grounded in visual evidence.
Existing token-level grounding methods primarily act at an object-level (i.e., via noun tokens), focusing on \emph{what} is in the image.
%
For instance, early works exploit raw attention weights that provide a readily available grounding signal, however they may reflect contextual or generic visual dependencies rather than token-specific grounding evidence~\citep{jain2019attention,serrano2019attention}.
Activation-based and gradient-based methods~\citep{wang2020scorecam,li2025tam,selvaraju2017gradcam} have also been adapted to MLLMs for token-level visual attribution, but they struggle due to the contextual dependence of subsequently generated tokens, and evaluations are largely limited to object nouns.
Current literature leaves a major blind spot in visual grounding, which we term the \emph{what-to-which} problem---the shift from object-level to instance-level grounding, a pivotal but still uncovered challenge in multi-instance applications and cluttered scene understanding. 
In fact, given the prompt ``the spotted dog sitting on the right'' and an image of dogs, noun grounding may highlight a ``dog'' in the image, but it is not guaranteed it is the intended referent. 
This ambiguity is solved at language level by additional details, such as \virg{spotted}, \virg{right}, and \virg{sitting}, but established grounding methods systematically discard these modifier tokens.
Recently, contextual-embedding methods~\citep{phukan2025beyond} show that contextual MLLM representations can support modifier grounding, yet they perform the analysis at the level of answer representations rather than at that of individual generated tokens.
Differently, grounding-oriented fine-tuning~\citep{chen2023shikra,peng2023kosmos2,zhang2023gpt4roi,zhang2023llavagrounding,ma2024groma} manages to improve localization by changing the model itself, neglecting to isolate grounding already present in its frozen representations.

Therefore, the \emph{what-to-which} problem remains unsolved at the generated-token level. 
We address this gap by combining two distinct but complementary objectives: first, by expanding the analysis beyond nouns to include instance-discriminative modifiers, we target the specific tokens required to resolve the referential ambiguity; second, by keeping the underlying model strictly frozen, we frame our analysis as a pure interpretability effort, aiming to uncover the grounding mechanisms already embedded in its internal representations.
This motivates our central question: \textit{how do frozen MLLMs visually ground the what-to-which problem across their generated tokens?}
To address this question, we introduce {OTTER} ({O}ptimal {T}ransport for {T}oken-level {E}xpression-to-{R}egion alignment), a lightweight supervised probe over frozen MLLM representations.
We approach token-level grounding as a matching problem between generated tokens and visual regions.
In particular, we train OTTER on Referring Instance Segmentation (RIS), aiming to identify more compatible token--region pairs and exclude those token or region with less meaningful visual match.
OTTER models this structure through entropy-regularized Optimal Transport (OT), producing token-level grounding maps without modifying the underlying MLLM.
Importantly, since accurate localization alone would not establish that modifier tokens carry instance-specific visual information,  
We design a controlled grounding protocol to test whether modifiers contribute grounding beyond object identity, to rule out simpler explanations for the recovered signal, and to assess if it extends beyond the controlled setting.

Our main contributions are the following:

\begin{itemize}[leftmargin=*,nosep]
    
    \item {We study the \emph{what-to-which} problem and propose a controlled grounding protocol.} We investigate whether token representations encode information that identifies a particular object instance beyond category-level object identity. To this end, our protocol rigorously separates instance-specific grounding from generic contextual, category-level, and supervision-induced signals.
    
    \item 
    To address the \emph{what-to-which} problem, we reframe it under the lens of Optimal Transport (OT). 
    We propose OTTER, a lightweight supervised probe that aligns generated tokens with visual regions through entropy-regularized OT, while keeping the underlying MLLM frozen.
    
    \item 
    Through controlled and out-of-distribution experiments, we show that instance-relevant visual information is recoverable from frozen MLLM representations and is distributed across contextualized token states. We demonstrate that this signal is particularly strong for spatial modifiers, but it is not exclusive to modifiers or object nouns.
\end{itemize}

The remainder of the paper is organized as follows. We outline the relevant literature that inspired and influenced our work in \Cref{sec:related}. 
Then, we describe the protocol used to extract and analyse token-visual mapping in \Cref{sec:protocols}. 
In \Cref{sec:method}, we describe the formulation of the OTTER module. 
Finally, we provide details on the experimental set up and results in \Cref{sec:exp_setup} and \ref{sec:results}. 
\section{Related Work}
\label{sec:related}

\paragraph{Token-level visual grounding for MLLMs.}
Recent work has extended visual grounding from image classifiers to autoregressive multimodal generation.
Token Activation Maps (TAM)~\citep{li2025tam} summarize contextual interference from preceding generated tokens and produce token-level visual maps without additional localization supervision in a zero-shot manner.
Other training-free or black-box approaches derive explanations from attention, activation maps, gradients, or input perturbations, such as LVLM-Interpret~\citep{stan2024lvlminterpret}, LLaVA-CAM~\citep{zhang2025redundancy}, and EAGLE~\citep{chen2025eagle}. 
These methods primarily evaluate generic word- or object-level grounding and faithfulness. 
OTTER addresses a complementary question: whether instance-discriminative modifiers, such as spatial terms and attributes, can be exploited as a distinct token category. 
Unlike the above methods, OTTER learns a lightweight grounding geometry from RIS supervision while keeping the underlying MLLM frozen.

\paragraph{Visual grounding from frozen MLLM features.}
A related line of work investigates whether visual grounding information can be decoded directly from frozen MLLM representations. 
Logit-lens-style approaches project internal representations to localize visually grounded concepts~\citep{nostalgebraist2020logitlens,jiang2024interpreting,neo2025interpreting}. 
Differently,~\citet{phukan2025beyond} use contextual mid-layer embeddings to show that internal representations contain grounding information beyond object nouns, 
at the answer level.
While sharing this premise, OTTER goes further by decoding signals at a token level. 
In fact, it learns a structured token--patch geometry yielding compact, token‑specific grounding maps which disentangle modifiers from head nouns and validate the functional relevance.

\paragraph{Supervised visual grounding and referring segmentation.}
Grounding-oriented MLLMs---e.g., Shikra~\citep{chen2023shikra}, Kosmos-2~\citep{peng2023kosmos2}, LLaVA-Grounding~\citep{zhang2023llavagrounding}, and Groma~\citep{ma2024groma}---train the model to predict localized outputs such as boxes or masks. 
RIS approaches, instead, use referring-expression comprehension and segmentation to localize the object specified by a natural-language expression~\citep{kazemzadeh2014referit,mao2016generation,yu2016modeling}.
OTTER is neither a grounding fine-tuning method nor a RIS segmenter. 
In fact, it leverages a frozen MLLM and uses RIS masks only as supervision for learning a modifier grounding probe over its existing token and visual representations. 

\paragraph{Optimal Transport for token--visual alignment.}
Optimal Transport addresses the problem of moving mass between two distributions at minimal cost.
In particular, Kantorovich's formulation \citep{kantorovich1942translocation} seeks an optimal transport plan: a coupling that specifies how mass from one distribution is matched to the other, with the resulting optimal cost defining the Wasserstein distance.
OTTER leverages an entropy-regularized relaxation of this problem given by the Sinkhorn distance \citep{cuturi2013sinkhorn}, aligning the mass of tokens with that of visual patches.
The regularization introduces the flexibility to assign non-groundable words or irrelevant patches to so-called \textit{dustbins}.

\section{Evaluation Protocol and Controls}
\label{sec:protocols}
To ground the \emph{what-to-which} problem, we develop a comprehensive protocol structured as seven guiding questions.
We first establish controlled instance-level grounding and analyze where the corresponding signal is represented (Q1--Q3). 
We then test the role of the learned token--visual mapping and the relevance of the recovered regions to generation (Q4--Q5), and finally assess whether the learned grounding generalizes to free-generation and across datasets (Q6--Q7). 

\paragraph{Q1. Does the full phrase improve instance-level grounding over the object noun alone?}
Free-generation grounding does not establish whether the linguistic information that specifies an instance improves grounding beyond the object category itself. 
We therefore use RefCOCO as a controlled setting and prompt the MLLM to generate either the full referring phrase, including attributes and modifiers, or only the object category noun. The resulting maps are compared with the ground-truth mask using pointing, Energy-in-GT, Soft-IoU, thresholded IoU, and entropy.

\paragraph{Q2. Which token types carry instance-discriminative grounding information?}
A phrase-level advantage does not reveal which linguistic components carry the instance-discriminative signal. We therefore introduce RefCOCO-MOD, a token-level partition of the RefCOCO validation set that groups generated tokens into spatial modifiers, attributes, relational/action terms, object nouns, and function words. We run TAM and OTTER 
on the same generated phrases and tokenization, and each token-level map is compared with the referred-instance mask using Soft-IoU.

\paragraph{Q3. Is referential information modifier-specific, or distributed across contextualized tokens?}
We test whether instance-specific grounding is confined to explicit modifiers or shared across representations. 
First, using RefCOCO counterfactual pairs (images with distinct same-category instances and different referring expressions), we use target-versus-distractor margins and switch consistency to test if modifier and noun maps correctly align with their respective targets. 
For a stricter control, we isolate a \emph{left}/\emph{right} subset where paired expressions share the noun and differ solely in opposing spatial terms.

Next, we test visual dependence by extracting token representations under correct, blank, or unrelated images, while grounding them against the original image patches. 
We also apply two text perturbations: \emph{scrambled} (averaged over five random word-order permutations) and \emph{isolated} (removing surrounding context). 
For token $i$ under control condition $\small\textbf{\texttt{c}}$ and in-context condition $\small\textbf{\texttt{ctx}}$, we define the retention ratio and its noun-normalized group version as
\begin{equation}
    \label{eq:retention_ratio}
    \small
    r_i^{(c)}=\frac{m_i^{(c)}}{m_i^{(\mathrm{ctx})}} \quad \text{and} \quad
    R_g^{(c)}=\frac{\operatorname{median}_{i \in g} r_i^{(c)}}{\operatorname{median}_{i \in \mathrm{noun}} r_i^{(c)}},
\end{equation}
where $m$ is the grounding metric. $R_g^{(c)}<1$ indicates greater context-dependence than the object noun.
Finally, a free-generation control evaluates cases where the object noun is generated before its spatial modifier. Due to autoregressive causal masking, this tests whether instance-specific information is already present in the noun state before the explicit modifier is produced.

\paragraph{Q4. Does grounding depend on the learned token--visual mapping?}
Good localization performance alone does not clarify whether OTTER is using information carried by the textual representations rather than supervision alone to determine where each token is grounded. We therefore compare OTTER with two matched controls that preserve the same implementation 
while altering only the semantic mapping between text and visual features. Specifically, we use \emph{raw cosine} 
with the original Qwen representations, 
and \emph{shuffled projection} replacing the learned projection with a fixed random channel permutation.

\paragraph{Q5. Are the recovered regions relevant to the model's own generation?}
Overlap with a ground-truth mask does not by itself show that the highlighted regions matter to the model's generation. We therefore perform deletion and insertion perturbations, masking or revealing image regions according to the grounding map 
and measuring the resulting change in token log-likelihood. 
We compare OTTER with random, center, and attention baselines. Stronger deletion effects and faster insertion recovery indicate greater relevance to the model's prediction.

\paragraph{Q6. Does OTTER generalize to token-level grounding under free generation?}
We test whether OTTER generalizes beyond RefCOCO-derived supervision to unconstrained free generation on COCO Caption and OpenPSG. Following the TAM protocol, we use the frozen Qwen2.5-VL backbone and evaluate both a fixed ($0.5$) and a per-image Otsu threshold, reporting Obj-IoU, Func-IoU, F1-IoU, Precision, and Recall.

\paragraph{Q7. Does the learned grounding geometry transfer across datasets?}
Generalization under free captioning does not establish that the learned probe transfers to a different referring-expression distribution. 
We therefore apply a trained OTTER checkpoint to RefCOCOg without retraining, where expressions are longer and more descriptive than in RefCOCO. 
We evaluate grounding against the referred-instance masks both at the phrase level and separately for nouns, attributes, spatial modifiers, and relational/action terms. 
This complements Q6 by testing cross-dataset transfer within the referring-expression family rather than unconstrained free-generation generalization.
\section{Method}
\label{sec:method}
\begin{figure}[htbp]
    \centering
    \includegraphics[height=3.5cm]{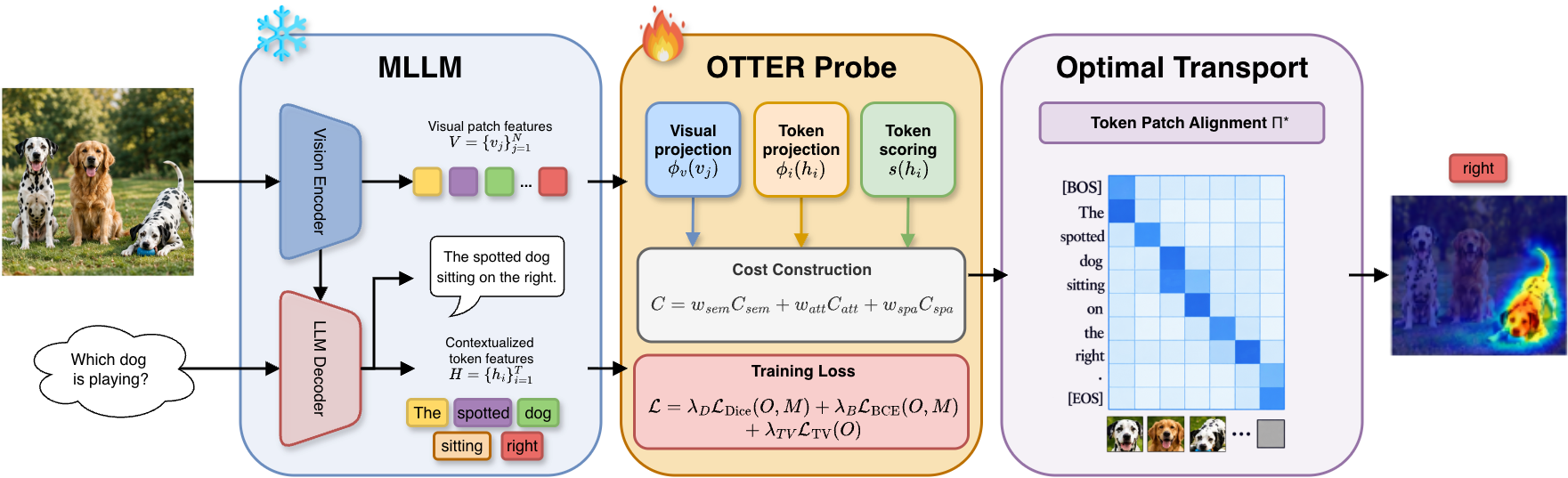}
        \caption{{Method Overview.} The frozen MLLM produces token and visual patch states, which the trainable OTTER probe projects into a shared space to build a token--patch cost matrix. 
        An entropy-regularized OT solver aligns tokens and visual patches, yielding a token-level grounding map.}
    \label{fig:otter_overview}
\end{figure}

OTTER is a lightweight head module that exploits features from the MLLM to provide token-region attribution for the generated modifiers. 
Starting from the contextual token states and visual patch states produced by a frozen MLLM, we train OTTER to learn an optimal projection of both representations into a shared embedding space under a supervised segmentation objective. 
To this purpose, we build a token-patch cost matrix accounting for semantic, attention and spatial features, and solve an entropy-regularized OT problem whose optimal plan is aggregated into a patch-level grounding map.
In what follows, we formalize the details of how OTTER learns from the frozen MLLM's internal representations to produce an optimal token-patch grounding map.
See~\Cref{alg:otterx} and~\Cref{fig:otter_overview} for algorithmic and graphical descriptions of the method.

\subsection{Problem Setup}
The frozen MLLM receives an image $I$ and a prompt $q$ as input, and generates a sequence of tokens
\begin{equation}
    y_{1:T} = (y_1,\ldots,y_T).
\end{equation} 
Additionally, the MLLM provides a contextual hidden state $h_i \in \mathbb{R}^{d_t}$ for each generated token $y_i$, and a set of visual patch states $V = \{v_j\}_{j=1}^{N}$ for each reference image, where $v_j \in \mathbb{R}^{d_v}$.
Our goal is to produce a token-level grounding map
\begin{equation}
    E_{i,j} \in[0,1]^{T\times N}
\end{equation}
where each $E_{i,j}$ measures how much patch $j$ supports token $i$.
For words composed of several tokens, we aggregate the token maps by a max aggregation, as in~\citet{li2025tam}.
\subsection{OTTER Head Training}
OTTER is a lightweight head trained on the frozen MLLM internal representations for RIS, specifically the
contextual token states $H=\{h_i\}_{i=1}^{T}$ and the visual states $V=\{v_j\}_{j=1}^{N}$.
Features are extracted from an intermediate layer of the MLLM selected via validation grounding diagnostics (see Appendix \ref{app:architecture_ablation}, Table \ref{tab:cross-backbone-layer-screen} and \ref{tab:mini-token-type-cross-backbone}).
The trainable head---comprising ${\approx}550$k learnable parameters compared 
to the billions 
of the frozen backbone---computes a token projection $\phi_t$, a visual projection $\phi_v$, and a token scoring module $s(h_i)$ that produces source masses as
\begin{equation}
    \label{eq:alpha_scores}
    \alpha_i = \operatorname{softmax}_i s(h_i).
\end{equation}
Specifically, these functions are implemented as linear layers with normalization.
During training we use RIS triples $(I,R,M)$, where $R$ is a referring expression and $M \in \{0,1\}^N$ is the target mask downsampled to the patch grid.
The trainable probe consists of two lightweight projections
\begin{equation}
    z_i^t = \frac{\phi_t(h_i)}{\|\phi_t(h_i)\|_2} \quad \text{and}
    \quad
    z_j^v = \frac{\phi_v(v_j)}{\|\phi_v(v_j)\|_2}
\end{equation}
which define a \emph{shared semantic space}. 
Thus we compute the cost between token \textit{i} and patch \textit{j} as
\begin{equation}
    C^{\mathrm{sem}}_{ij} = 1 - \langle z_i^t, z_j^v \rangle.
    \label{eq:semantic_cost}
\end{equation}

When the selected head uses attention and spatial terms, the implementation normalizes the attention cost per token and the spatial cost by its maximum value:
\begin{align}
    C^{\mathrm{att}}_{ij}
    &=
    \frac{-\log \tilde{A}_{ij}}{\frac{1}{N}\sum_k -\log \tilde{A}_{ik}},
    \qquad
    \tilde{A}_{ij} = \frac{A_{ij}}{\sum_k A_{ik}}, \\
    C^{\mathrm{spa}}_{ij}
    &=
    \frac{\|p_j - \mu_i\|_2^2}{\max_{i,j}\|p_j-\mu_i\|_2^2},
    \qquad
    \mu_i = \sum_j \tilde{A}_{ij} p_j,
\end{align}
where $p_j$ is the patch coordinate.
Therefore, the full non-gated cost is
\begin{equation}
    C_{ij}
    =
    w_s C^{\mathrm{sem}}_{ij}
    +
    w_a g^a_i C^{\mathrm{att}}_{ij}
    +
    w_p g^p_i C^{\mathrm{spa}}_{ij}
    \label{eq:gated_full_cost}
\end{equation}
where $w_s, w_a, w_p \in[0,1]$ are learnable weights and $g^a_i,g^p_i\in[0,1]$ are optional token-specific gates that can be learnt to modulate the attention and spatial terms.
In Appendix~\ref{app:architecture_ablation}, we present an architecture ablation by showing the results of different cost configurations.
Specifically, we 
include/exclude semantic, spatial and attention costs, as well as fix/learn adaptive-gated heads, weights and dustbins. 

\subsection{Token--Patch Optimal Transport}
From the token--patch cost matrix $C$, OTTER solves an entropy-regularized optimal transport problem.
To define the OT marginals $a$ and $b$, we first compute scores for both text tokens and visual patches using Eq.~\ref{eq:alpha_scores}.
We then augment both marginals with dustbin entries, allowing transport mass to be assigned to non-groundable text tokens and irrelevant visual patches.
Both marginals are rescaled to have the same total mass before solving OT. 
Let $a \in \Delta^{T+1}$ and $b \in \Delta^{N+1}$ denote the resulting source and target marginals, then the optimal plan is defined as
\begin{equation}
    \Pi^\star
    =
    \arg\min_{\Pi \in \mathcal{U}(a,b)}
    \langle \Pi,C \rangle
    +
    \varepsilon \sum_{ij} \Pi_{ij}(\log \Pi_{ij}-1),
    \label{eq:ot}
\end{equation}
where $\mathcal{U}(a,b)$ is the set of couplings with marginals $a,b$ and
$\varepsilon{=}0.05$ is the Sinkhorn temperature.
We solve the resulting problem with $80$ log-domain Sinkhorn iterations.
The patch evidence is obtained by summing the transported mass from text tokens to visual patches:
\begin{equation}
    O_j = \sum_{i=1}^{T} \Pi^\star_{ij}, \qquad j=1,\ldots,N.
\end{equation}
During head training, the transported patch mass is converted into the grid score map by multiplying by the number of visual patches and clamping to $[0,1]$.
Official threshold metrics resize this raw score map to image resolution without per-example min--max normalization.
We remark that, since the backbone is frozen, the probe learns how to decode grounding evidence from existing MLLM representations rather than modifying the model's generation behaviour.

\paragraph{Residual Semantic Expansion.}
Raw OT is precise in pointing but lacks coverage, whereas the semantic branch improves recall at the cost of being diffuse. 
We combine both these approaches in the OTTER-Residual configuration:
\begin{equation}
    E^{\mathrm{res}}_{i}
    =
    O_i + \beta S_t (1-O_i),
    \label{eq:residual}
\end{equation}
where $\beta \in [0,1]$ and $S_t$ is the normalized learned semantic evidence obtained from $1-C^{\mathrm{sem}}_{ij}$.
\paragraph{Grounding Masks Refinement.} 
We use a lightweight morphological post-processing refinement on thresholded maps: binary closing with $2$ iterations, hole filling, and removal of connected components smaller than $0.05\%$ of the image area. 
Ablation on other refinements is in Appendix~\ref{app:refinement_ablation}.
\subsection{Training Objective}
\label{sec:training_objective}
For a predicted score map $O$ and target mask $M$ on the visual grid, the training objective is 
\begin{equation}
    \mathcal{L}_{\mathrm{mask}}
    =
    \lambda_D \mathcal{L}_{\mathrm{Dice}}(O,M)
    +
    \lambda_B \mathcal{L}_{\mathrm{FocalBCE}}(O,M)
    +
    \lambda_{TV} \mathcal{L}_{\mathrm{TV}}(O).
    \label{eq:mask_loss}
\end{equation}
Here, $\mathcal{L}_{\mathrm{Dice}}$ promotes overlap between the predicted and target masks~\citep{milletari2016vnet}, $\mathcal{L}_{\mathrm{FocalBCE}}$ down-weights easy foreground/background predictions~\citep{lin2017focal}, and $\mathcal{L}_{\mathrm{TV}}$ regularizes local spatial variation~\citep{rudin1992nonlinear}.
\begin{algorithm}[b]
\caption{OTTER generated-token grounding}
\label{alg:otterx}
\small
\begin{algorithmic}[1]
\REQUIRE Image $I$, prompt $q$, frozen MLLM $f$, probe trained with Eq.(\ref{eq:mask_loss}), residual weight $\beta$.
\STATE Generate tokens $y_{1:T}$ with $f(I,q)$; store hidden states $H=\{h_i\}_{i=1}^T$; extract visual patches $V=\{v_j\}_{j=1}^{N}$.
\FOR{each generated token $t$}
    \STATE Project $h_i$ and $v_j$ into the learned token--region space.
    \STATE Compute the token--patch costs in Eq.(\ref{eq:gated_full_cost}): 
    semantic cost, (gated) attention and spatial costs.
    \STATE Add text and visual dustbins and solve the Sinkhorn OT problem in Eq.(\ref{eq:ot}).
    \STATE Obtain raw OT map $O_i$ from transported patch mass, rescaled by $N$ and clamped to $[0,1]$.
    \STATE Optionally compute residual map $E_i^{\mathrm{res}}=O_i+\beta S_t(1-O_i)$.
\ENDFOR
\RETURN Token-level maps $\{O_i\}$ and optional residual maps $\{E_i^{\mathrm{res}}\}$.
\end{algorithmic}
\end{algorithm}

\section{Experimental Setup}
\label{sec:exp_setup}

\paragraph{Methods.}
Our probing module is trained on hidden states and visual patch states extracted from a frozen Qwen2.5-VL model~\citep{bai2025qwen25vl}.
We additionally screen OTTER on three further backbones (Qwen3-VL-2B, InternVL3.5-2B, Gemma-3-4B), and the layer-selection protocol and results are reported in Appendix~\ref{app:architecture_ablation}.
Unless otherwise stated, OTTER denotes the adaptive-gated OT head with learned global cost weights and learned text/visual dustbin costs.
Each OTTER's head is optimized with Eq.(\ref{eq:mask_loss}), with $\lambda_D=1.0,\lambda_B=0.5$ and $\lambda_{TV}=0.02$.
Optimization uses AdamW with learning rate $2\cdot 10^{-4}$, weight decay $10^{-4}$, and gradient clipping at $1.0$.
Early stopping is decided based on the calibration set mIoU.
We calculate the mIoU for both fixed ($0.50$), and dynamic thresholds (selected on the calibration split).
We evaluate OTTER against the learned semantic branch, attention maps, center priors, and random maps.
Ablations on the full head-family, training-scale, and transport-operator are reported in Appendix~\ref{app:architecture_ablation}.

\paragraph{Datasets and evaluation settings.}
OTTER is trained on $2{,}500$ images from RefCOCO~\citep{yu2016modeling}, providing the referring expressions and instance masks, calibration, and controlled evaluation.
RefCOCO-MOD (see
~\Cref{sec:protocols}) contains $5{,}021$ spatial, $2{,}722$ attribute, and $516$ relational/action tokens, together with $15{,}881$ noun and $5,017$ function-word tokens used as controls.
COCO Caption~\citep{chen2015microsoft} ($5$
k images) and OpenPSG~\citep{zhou2024openpsg} ($3{,}176$ images) are used for evaluation under free-generation.
RefCOCOg is used only for cross-dataset transfer within the referring-expression family, where grounding is evaluated both at the phrase level and by generated token group.
The context-control analysis uses the full RefCOCO validation split, with $10{,}834$ expressions attempted and $8{,}407$ retained after removing single-word and modifier-free cases.

\paragraph{Counterfactual referent controls.}
Starting from the RefCOCO validation expressions used in the contextual analysis, we construct 2,971 natural counterfactual pairs spanning 1,238 images. 
Each pair contains distinct referred instances within the same image and object category, together with referring expressions that provide a discriminative modifier contrast. 
We also define a stricter subset of 217 clean left/right pairs over 187 images, requiring distinct target annotations, a shared clean object noun, opposing left/leftmost versus right/rightmost expressions, and no attribute modifiers. 

\paragraph{Visual and context perturbations.}
For each counterfactual pair, we evaluate noun and modifier maps under three visual conditions, extracting token hidden states after replacing the original (\emph{correct-}) image with either a blank (\emph{blank-}) or unrelated (\emph{wrong-}) image, while OTTER grounds those states against the original image patches, so that any drop reflects lost target-specific information rather than a different grounding target. We report target preference, target-versus-distractor margin, and bidirectional switch consistency for each condition. We further perturb the referring context itself, via \emph{scrambled} (five random word-order permutations) and \emph{isolated} (context removed) variants, reporting the retention ratio (see Eq.~(\ref{eq:retention_ratio})).

\paragraph{Faithfulness evaluation.}
The perturbation analysis uses generated phrases from the RefCOCO validation set images, expanded to the per-token unit level.
Deletion and Insertion use a fixed gray occluder.
Method differences are summarized with paired bootstrap confidence intervals using 1{,}000 resamples of the shared example IDs.

\paragraph{Transfer evaluation.}
RefCOCOg is evaluated without retraining on RefCOCOg data, using the cleanest transfer operating point (the fixed full-OT) checkpoint, rather than the adaptive-gated checkpoint used as the selected primary architecture elsewhere in the paper.

\section{Results}
\label{sec:results}

We organize the results following the evaluation protocol in
Section~\ref{sec:protocols}. 
We first establish controlled instance-level grounding and examine where the corresponding signal is represented (Q1--Q3). We then test the role of the learned token--visual mapping and the relevance of the recovered regions to generation (Q4--Q5). 
Finally, we evaluate generalization to unconstrained free generation and cross-dataset transfer (Q6--Q7).


\subsection{Controlled Instance-Level Grounding (Q1--Q2)}
\label{sec:controlled_modifier_results}

\paragraph{Phrase versus noun-only generation.}
Table~\ref{tab:refcoco_controlled_generated_token_grounding} shows that
grounding the full referring expression is more informative than grounding
the object noun alone. This establishes that the linguistic context
specifying an instance provides useful grounding information beyond the
object category. Importantly, this phrase-level comparison does not
establish that the additional information is localized specifically in
modifier representations; we address that distinction at the token level
in Q2--Q3.

\paragraph{Grounding across token categories.}
RefCOCO-MOD allows us to ask where instance-level visual information can be
decoded within the generated expression. As shown in
Table~\ref{tab:refcoco_controlled_generated_token_grounding_tam_comparison},
the learned probe recovers grounding information across several linguistic
roles, with particularly strong results for object nouns and spatial
modifiers, followed by attributes. Relational/action terms provide a
weaker but consistent signal. Function words also contain non-trivial
target information under the supervised probe. We do not interpret this
as independent lexical visual semantics; rather, it suggests that
referential information may be distributed through contextualized token
representations. This observation motivates Q3, where we test directly
whether the referred instance is encoded specifically by modifiers or
shared across tokens.


\begin{table}[htpb]
\centering

\begin{minipage}[t]{0.49\linewidth}
\vspace{0pt}
\centering
\scriptsize
\setlength{\tabcolsep}{1.5pt}
\renewcommand{\arraystretch}{0.95}

\captionof{table}{
Controlled RefCOCO grounding.
}
\label{tab:refcoco_controlled_generated_token_grounding}

\vspace{0.8mm}

\begin{tabular}{@{}llcccc@{}}
\toprule
Protocol & Method
& Energy $\uparrow$
& Soft-IoU $\uparrow$
& IoU@.4 $\uparrow$
& Ent. $\downarrow$ \\
\midrule

\multirow{3}{*}{Phrase}
& Semantic
& 0.170 & 0.156 & 0.177 & 0.993 \\

& OTTER
& \textbf{0.405}
& \textbf{0.264}
& 0.332
& \textbf{0.891} \\

& OTTER-R $\beta{=}0.3$
& 0.254
& 0.205
& \textbf{0.348}
& 0.974 \\

\midrule

\multirow{3}{*}{Noun-only}
& Semantic
& 0.162 & 0.147 & 0.168 & 0.993 \\

& OTTER
& \textbf{0.310}
& \textbf{0.189}
& 0.229
& \textbf{0.907} \\

& OTTER-R $\beta{=}0.4$
& 0.205
& 0.166
& \textbf{0.264}
& 0.983 \\

\bottomrule
\end{tabular}

\end{minipage}
\hfill
\begin{minipage}[t]{0.49\linewidth}
\vspace{0pt}
\centering
\scriptsize
\setlength{\tabcolsep}{1.5pt}
\renewcommand{\arraystretch}{0.95}

\captionof{table}{
RefCOCO-MOD Soft-IoU comparison.\\
}
\label{tab:refcoco_controlled_generated_token_grounding_tam_comparison}

\vspace{0.8mm}

\begin{tabular}{@{}lrrrl@{}}
\toprule
Token group & $n$ & TAM & OTTER & $\Delta$ [95\% CI] \\
\midrule

Phrase
& 10834 & 0.118 & \textbf{0.342}
& $+0.224\ [0.221,\,0.228]$ \\

Nouns
& 15881 & 0.106 & \textbf{0.310}
& $+0.204\ [0.201,\,0.207]$ \\

Spatial
& 5021 & 0.061 & \textbf{0.304}
& $+0.243\ [0.238,\,0.248]$ \\

Attributes
& 2722 & 0.119 & \textbf{0.285}
& $+0.167\ [0.161,\,0.173]$ \\

Rel./action
& 516 & 0.077 & \textbf{0.172}
& $+0.095\ [0.082,\,0.110]$ \\

Function
& 5017 & 0.060 & \textbf{0.244}
& $+0.184\ [0.179,\,0.189]$ \\

\bottomrule
\end{tabular}

\end{minipage}

\end{table}


\subsection{Distributed Referential Grounding and Context Dependence (Q3)}
\label{sec:distributed_grounding}

Q1--Q2 show that modifier states contain instance-level visual information,
but not whether this information is specific to modifiers. We therefore
test how referential information is distributed across contextualized
tokens.

\paragraph{Counterfactual referent switching.}
On natural same-image counterfactual pairs, modifier maps reliably follow
the intended same-category referent, but contextualized noun maps do so
just as reliably and can be even more selective
(Table~\ref{tab:counterfactual}, top). The same pattern holds in the
stricter left/right subset. Thus, the linguistic distinction between
\emph{what} and \emph{which} does not correspond to a strict
noun--modifier separation in the internal representations: instance
identity is distributed across contextualized token states. Decomposing
by modifier type (Table~\ref{tab:counterfactual}, bottom), this pattern is
clearest for spatial expressions; attribute contrasts are broadly
consistent, while relational/action pairs are too sparse for a reliable
conclusion.

\begin{table*}[htpb]
\centering

\begin{minipage}[t]{0.5\textwidth}
\vspace{0pt}
\captionof{table}{
Counterfactual referent grounding.
}
\label{tab:counterfactual}
\centering

{\fontsize{6.0}{6.6}\selectfont
\setlength{\tabcolsep}{2.2pt}
\renewcommand{\arraystretch}{0.92}

\begin{tabular}{@{}llrcccc@{}}
\toprule
& & &
\multicolumn{2}{c}{Target pref. $\uparrow$} &
\multicolumn{2}{c}{Gap $\uparrow$}
\\
\cmidrule(lr){4-5}
\cmidrule(lr){6-7}

Group &
Condition &
Pairs &
Mod. &
Noun &
Mod. &
Noun
\\
\midrule

\multirow{3}{*}{All natural}
& Correct & 2971 & 0.776 & 0.779 & 0.203 & 0.242
\\
& Blank   & 2971 & 0.703 & 0.705 & 0.112 & 0.137
\\
& Wrong   & 2971 & 0.714 & 0.717 & 0.121 & 0.146
\\

\midrule

\multirow{3}{*}{Clean L/R}
& Correct & 217 & 0.772 & 0.795 & 0.220 & 0.293
\\
& Blank   & 217 & 0.677 & 0.744 & 0.115 & 0.193
\\
& Wrong   & 217 & 0.705 & 0.749 & 0.131 & 0.204
\\

\midrule

\multirow{3}{*}{By type}
& Spatial    & 2578 & 0.775 & 0.778 & 0.202 & 0.237
\\
& Attribute  & 173  & 0.786 & 0.769 & 0.197 & 0.244
\\
& Rel./act.  & 3    & 0.833 & 0.833 & 0.306 & 0.219
\\

\bottomrule
\end{tabular}
}

\vspace{1mm}

\captionsetup{
    font=footnotesize,
    justification=centering,
    singlelinecheck=false,
    skip=2pt
}

\end{minipage}
\hspace{0.02\textwidth}
\begin{minipage}[t]{0.40\textwidth}
\vspace{0pt}
\captionof{table}{
Deletion AUC by token group.
}
\label{tab:faithfulness_stratified}
\centering

{\fontsize{6.8}{7.4}\selectfont
\setlength{\tabcolsep}{2.7pt}
\renewcommand{\arraystretch}{1.02}

\begin{tabular}{@{}lccccc@{}}
\toprule
Method &
Phrase &
Noun &
Mod. &
Spatial &
Attr.
\\
\midrule

Random
& 0.310
& 0.305
& 0.182
& 0.151
& 0.226
\\

Attention
& 0.386
& 0.383
& 0.202
& 0.147
& 0.281
\\

Center
& 0.317
& 0.318
& 0.194
& 0.147
& 0.266
\\

Semantic
& 0.541
& 0.536
& 0.320
& 0.227
& 0.450
\\

\midrule
OTTER           & 0.543 & 0.535 & 0.322 & \textbf{0.234} & 0.445 \\

\textbf{OTTER-R}
& \textbf{0.546}
& \textbf{0.545}
& \textbf{0.324}
& 0.231
& \textbf{0.456}
\\

\bottomrule
\end{tabular}
}

\vspace{1mm}

\captionsetup{
    font=footnotesize,
    justification=centering,
    singlelinecheck=false,
    skip=2pt
}

\end{minipage}

\end{table*}
















\paragraph{Visual and contextual dependence.}
Replacing the correct image with a blank or unrelated one weakens
referential selectivity for both nouns and modifiers, showing that the
shared signal is not purely linguistic. Context perturbations further
reveal that this distribution is not uniform: function-word grounding
depends strongly on the surrounding expression, whereas spatial modifiers
remain comparatively robust (Table~\ref{tab:matched_controls}, left).
Referential information is therefore shared across tokens, but different
token classes depend on context to different degrees.

\paragraph{Referential information before the modifier.}
In free generation we isolate expressions where the noun is produced
before the disambiguating spatial modifier. Even in this causal setting,
noun states already contain significant information about the intended
instance, while the subsequently generated modifier is more discriminative.
This suggests that the \emph{what-to-which} transition can begin before
the explicit modifier is generated and become sharper as the referring
expression unfolds.


\subsection{Matched Probe Controls (Q4)}
\label{sec:control_probes}

The matched controls in Table~\ref{tab:matched_controls}, right, produce
a substantial performance collapse when the learned token--visual
projection is replaced by either raw feature similarity or a shuffled
mapping. Thus, the observed grounding cannot be attributed simply to RIS
supervision, the OT operator, or an immediately accessible cosine geometry
in the frozen features. Instead, the results support a more specific claim:
frozen Qwen representations contain referential information that can be
recovered by a lightweight learned projection. Since that projection is
itself trained with RIS supervision, we do not interpret this as evidence
for a fully pre-existing grounding geometry.

\begin{table}[htpb]
\centering
\renewcommand{\arraystretch}{1.05}

\caption{
Context dependence and matched probe controls.
(a) normalized Energy-in-GT retention under context perturbation;
(b) matched projection controls on RefCOCO-val under the same
OT/RIS training.
}
\label{tab:matched_controls}

\vspace{1.5mm}

\begin{minipage}[t]{0.55\linewidth}
\vspace{0pt}
\centering

\scriptsize 
\setlength{\tabcolsep}{4.5pt}

\vspace{-3mm}

\textbf{(a) Context perturbation}

\vspace{1mm}

\begin{tabular}{@{}lcc@{}}
\toprule
Token group & Scrambled & Isolated \\
\midrule

Nouns
& $1.000\,[0.99,1.01]$
& $1.000\,[0.98,1.02]$ \\

Spatial
& $0.982\,[0.97,0.99]$
& $\mathbf{1.077\,[1.05,1.10]}$ \\

Attribute
& $0.942\,[0.93,0.96]$
& $0.914\,[0.87,0.96]$ \\

Function
& $0.939\,[0.93,0.95]$
& $0.597\,[0.57,0.63]$ \\

\bottomrule
\end{tabular}

\end{minipage}
\hfill
\begin{minipage}[t]{0.42\linewidth}
\vspace{0pt}
\centering

\scriptsize 
\setlength{\tabcolsep}{4.5pt}
\vspace{-3mm}
\textbf{(b) Projection controls}

\vspace{1mm}

\begin{tabular}{@{}lcc@{}}
\toprule
Probe
&
\shortstack{Learned\\projection?}
&
\shortstack{Calibrated\\mIoU $\uparrow$} \\
\midrule

OTTER
& yes
& \textbf{0.458} \\

Raw cosine
& no
& 0.130 \\

Shuffled projection
& no
& 0.135 \\

\bottomrule
\end{tabular}

\end{minipage}

\end{table}

\subsection{Model Relevance under Perturbation (Q5)}
\label{sec:faithfulness_results}

Deletion experiments provide complementary evidence that the recovered
regions are relevant to the model's own token generation. Across token
groups, removing regions prioritized by OTTER (mainly in the Residual counterpart) has a stronger effect than
removing random, center-selected, or attention-selected regions
(Table~\ref{tab:faithfulness_stratified}). The learned semantic branch
behaves similarly under this intervention, indicating that the main
distinction introduced by OT lies in the spatial structure and
concentration of the recovered maps rather than in their perturbation
sensitivity. We therefore interpret deletion as evidence of model
relevance, rather than as a complete causal account of token generation.















\subsection{Generalization beyond the Controlled Setting (Q6--Q7)}
\label{sec:generalization_results}

\paragraph{Free generation (Q6).}
Table~\ref{tab:refcoco_openpsg_free_generation_tam_protocol} evaluates
OTTER on unconstrained generated text grounding on COCO Caption and
OpenPSG. Because TAM is zero-shot whereas OTTER is RIS-supervised, we do
not interpret the comparison as an intrinsic ranking of attribution
methods. Rather, the results show that the grounding learned from
RefCOCO-derived supervision remains useful when the model generates
unconstrained captions outside the controlled referring-expression setup.
\Cref{fig:qualitative_token_grounding} illustrates this qualitatively: for four free-generation examples, OTTER produces more concentrated, token-aligned maps than TAM, attention, the semantic branch, and the center/random baselines.

\paragraph{Cross-dataset transfer (Q7).}
Table~\ref{tab:transfer_refcocog} shows that the learned token--region
mapping remains effective when transferred to RefCOCOg without retraining.
The improvement extends beyond phrase-level grounding to nouns,
attributes, spatial expressions, and relational/action terms. Together,
Q6 and Q7 indicate that OTTER is not restricted to the specific RefCOCO
expression distribution used for probe supervision: the learned grounding
geometry transfers both to unconstrained generation and to a different
referring-expression dataset.


\begin{table}[!t]
\centering

\begin{minipage}[t]{0.63\linewidth}
\vspace{0pt}
\centering
\scriptsize
\setlength{\tabcolsep}{1.3pt}
\renewcommand{\arraystretch}{0.95}

\captionof{table}{
Free-generation grounding on COCO Caption and OpenPSG. \textbf{Best} in bold;
\underline{second-best} underlined.
}
\label{tab:refcoco_openpsg_free_generation_tam_protocol}

\vspace{0.8mm}

\begin{tabular}{@{}llcccccc@{}}
\toprule
Data
& Method
& Thr.
& Obj.~$\uparrow$
& Func.
& F1~$\uparrow$
& Prec.~$\uparrow$
& Rec.~$\uparrow$ \\
\midrule

\multirow{4}{*}{COCO}

& OTTER
& Otsu
& 0.193
& 0.929
& 0.320
& \textbf{0.471}
& 0.290 \\

& OTTER-R ($\beta{=}0.5$)
& $0.5$
& \underline{0.273}
& 0.864
& \textbf{0.415}
& 0.424
& 0.504 \\

& TAM
& Otsu
& \textbf{0.278}
& 0.778
& \underline{0.409}
& 0.404
& \textbf{0.668} \\

& TAM
& $0.5$
& 0.255
& 0.869
& 0.395
& \underline{0.450}
& \underline{0.539} \\

\midrule

\multirow{4}{*}{OpenPSG}

& OTTER
& Otsu
& 0.210
& 0.934
& 0.343
& \textbf{0.479}
& 0.300 \\

& OTTER-R ($\beta{=}0.5$)
& $0.5$
& \textbf{0.284}
& 0.863
& \textbf{0.427}
& 0.425
& \underline{0.524} \\

& TAM
& Otsu
& \underline{0.258}
& 0.784
& \underline{0.388}
& 0.385
& \textbf{0.644} \\

& TAM
& $0.5$
& 0.235
& 0.873
& 0.370
& \underline{0.428}
& 0.515 \\

\bottomrule
\end{tabular}

\end{minipage}
\hfill
\begin{minipage}[t]{0.36\linewidth}
\vspace{0pt}
\centering
\scriptsize
\setlength{\tabcolsep}{2.4pt}
\renewcommand{\arraystretch}{0.95}

\captionof{table}{
RefCOCOg transfer without retraining. Thresholded mIoU. \textbf{Best} in bold.
}
\label{tab:transfer_refcocog}

\vspace{0.8mm}

\begin{tabular}{@{}lcc@{}}
\toprule
Evaluation group & TAM & OTTER \\
\midrule

Phrase
& 0.171
& \textbf{0.372} \\

Nouns
& 0.134
& \textbf{0.332} \\

Attributes
& 0.152
& \textbf{0.328} \\

Spatial modifiers
& 0.053
& \textbf{0.328} \\

Rel./action
& 0.077
& \textbf{0.208} \\

\bottomrule
\end{tabular}

\end{minipage}

\end{table}


%
\definecolor{tokctx}{HTML}{F4C25B} 
\definecolor{tokattr}{HTML}{9973A8} 
\definecolor{toknoun}{HTML}{8EC06C} 
\definecolor{tokact}{HTML}{DE8F4E} 
\definecolor{tokspat}{HTML}{D9534F} 

\newcommand{\tokctx}[1]{\textcolor{tokctx}{#1}}
\newcommand{\tokattr}[1]{\textcolor{tokattr}{#1}}
\newcommand{\toknoun}[1]{\textcolor{toknoun}{#1}}
\newcommand{\tokact}[1]{\textcolor{tokact}{#1}}
\newcommand{\tokspat}[1]{\textcolor{tokspat}{#1}}

\newcommand{\otterprefix}{OTTER}

\newcommand{\tokbox}[1]{%
\includegraphics[
    width=0.11\textwidth,
    height=0.11\textwidth
]{#1}%
}

\newcommand{\tokboxsmall}[1]{%
\includegraphics[
    width=0.08\textwidth,
    height=0.08\textwidth
]{#1}%
}

\newcommand{\methodlabel}[1]{\tiny{#1}}
\newcommand{\tokenlabel}[1]{\tiny{#1}}

\begin{figure*}[htpb]
\centering
\setlength{\tabcolsep}{1pt}
\renewcommand{\arraystretch}{0.92}
\scriptsize

\begin{tabular}{c c c c c c c c c c c c}
& \multicolumn{3}{c}{\textit{Girl in purple}} 
& \multicolumn{2}{c}{\textit{Bottom cat}} 
& \multicolumn{3}{c}{\textit{Guy standing playing Wii Fit}} 
& \multicolumn{3}{c}{\textit{Cat on top}} \\[1mm]
& \textit{\toknoun{girl}} & \textit{\tokctx{in}} & \textit{\tokattr{purple}} 
& \textit{\tokspat{bottom}} & \textit{\toknoun{cat}} 
& \textit{\toknoun{guy}} & \textit{\tokact{standing}} & \textit{\tokact{playing}} 
& \textit{\toknoun{cat}} & \textit{\tokctx{on}} & \textit{\tokspat{top}} \\[1mm]

\rotatebox{90}{{\textbf{OTTER}}} & 
\tokboxsmall{images/best/01_idx046_siou0.718_girl_in_purple/\otterprefix_tok_girl_pos00_overlay.png} & 
\tokboxsmall{images/best/01_idx046_siou0.718_girl_in_purple/\otterprefix_tok_in_pos01_overlay.png} & 
\tokboxsmall{images/best/01_idx046_siou0.718_girl_in_purple/\otterprefix_tok_purple_pos02_overlay.png} & 
\tokboxsmall{images/best/02_idx038_siou0.607_bottom_cat/\otterprefix_tok_bottom_pos00_overlay.png} & 
\tokboxsmall{images/best/02_idx038_siou0.607_bottom_cat/\otterprefix_tok_cat_pos01_overlay.png} & 
\tokboxsmall{images/best/03_idx041_siou0.601_guy_standing_playing_wiifit/\otterprefix_tok_guy_pos00_overlay.png} & 
\tokboxsmall{images/best/03_idx041_siou0.601_guy_standing_playing_wiifit/\otterprefix_tok_standing_pos01_overlay.png} & 
\tokboxsmall{images/best/03_idx041_siou0.601_guy_standing_playing_wiifit/\otterprefix_tok_playing_pos02_overlay.png} & 
\tokboxsmall{images/best/08_idx033_siou0.426_cat_on_top/\otterprefix_tok_cat_pos00_overlay.png} & 
\tokboxsmall{images/best/08_idx033_siou0.426_cat_on_top/\otterprefix_tok_on_pos01_overlay.png} & 
\tokboxsmall{images/best/08_idx033_siou0.426_cat_on_top/\otterprefix_tok_top_pos02_overlay.png} \\

\rotatebox{90}{{TAM}} & 
\tokboxsmall{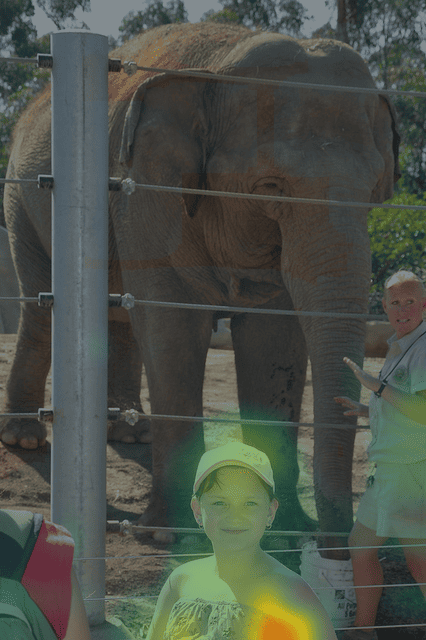} & 
\tokboxsmall{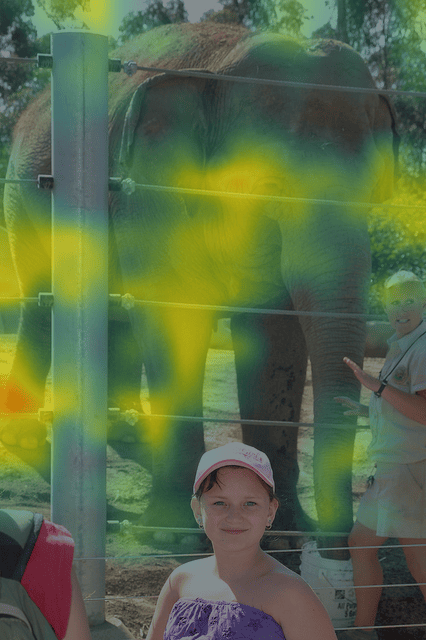} & 
\tokboxsmall{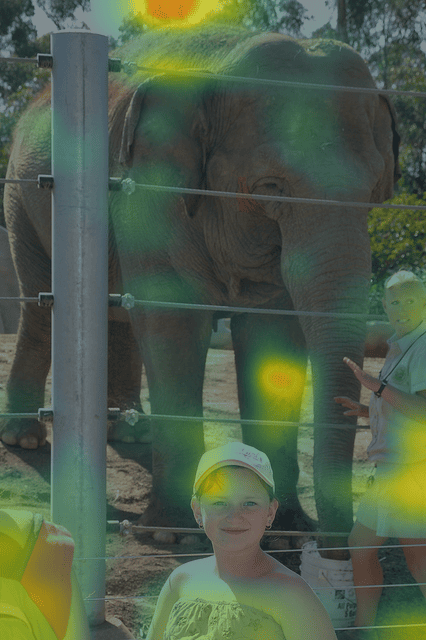} & 
\tokboxsmall{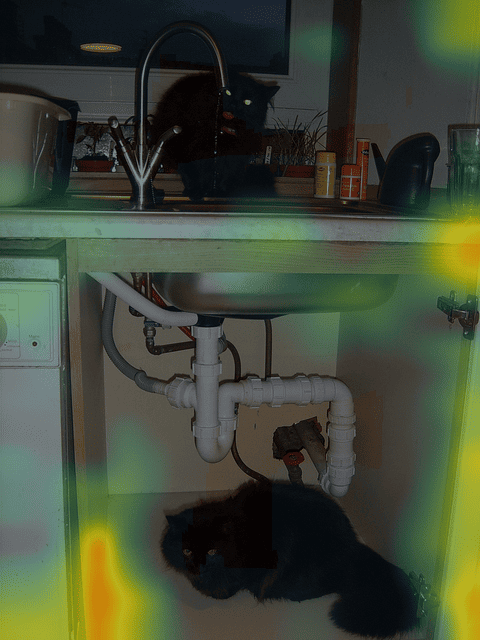} & 
\tokboxsmall{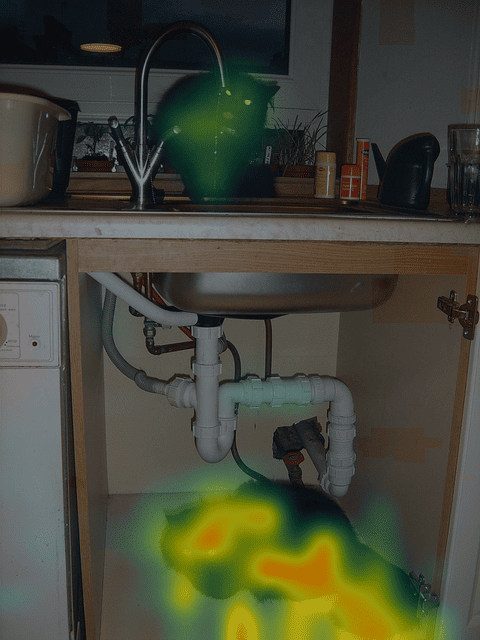} & 
\tokboxsmall{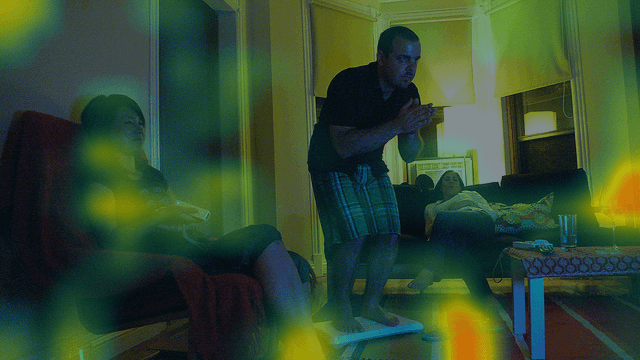} & 
\tokboxsmall{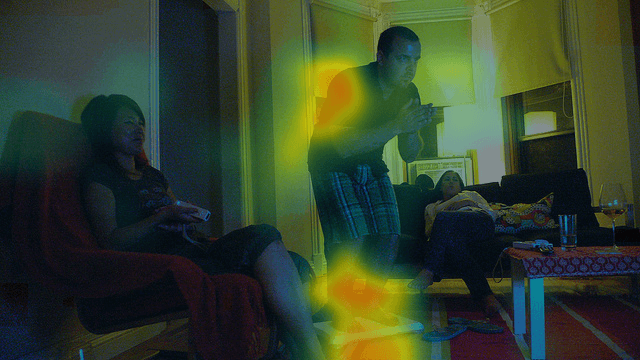} & 
\tokboxsmall{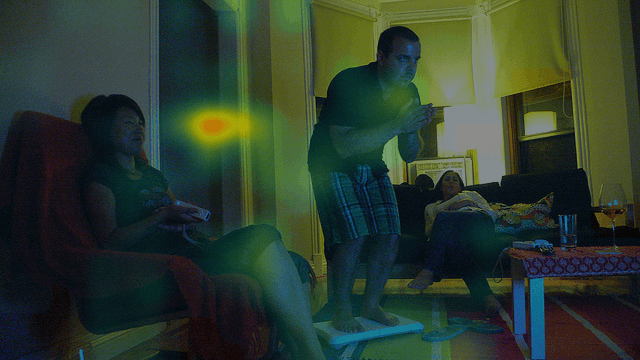} & 
\tokboxsmall{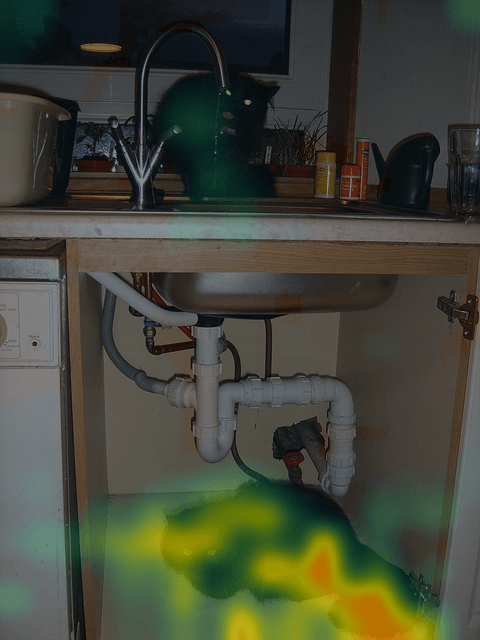} & 
\tokboxsmall{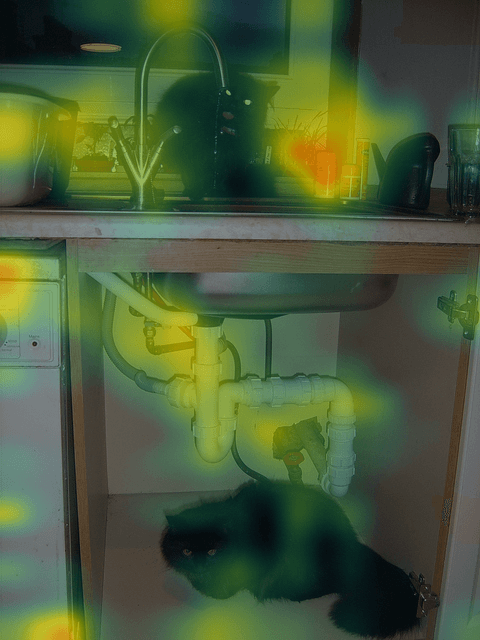} & 
\tokboxsmall{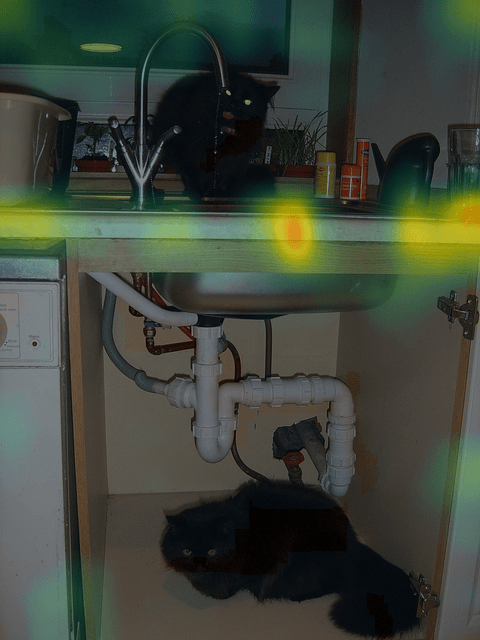} \\

\rotatebox{90}{{Attention}} & 
\tokboxsmall{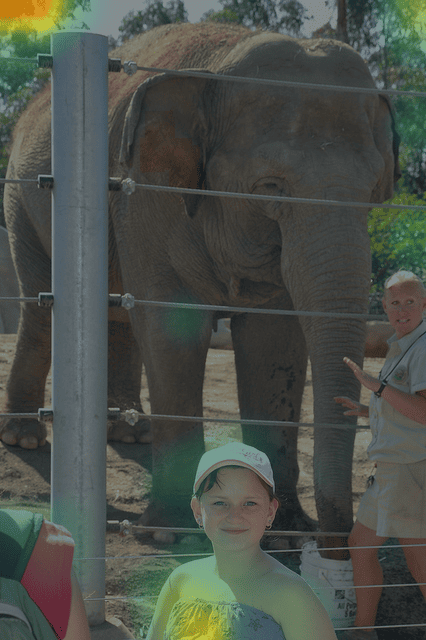} & 
\tokboxsmall{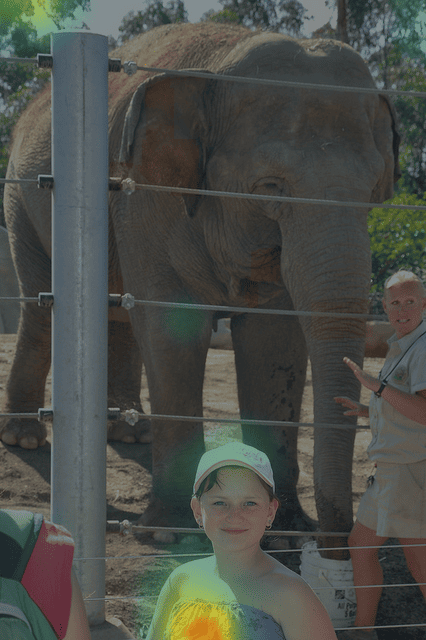} & 
\tokboxsmall{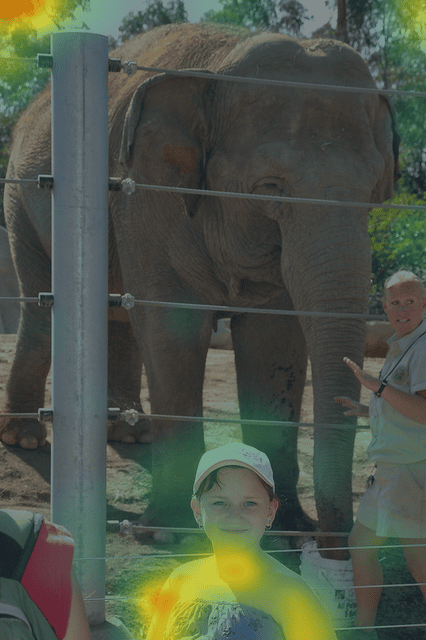} & 
\tokboxsmall{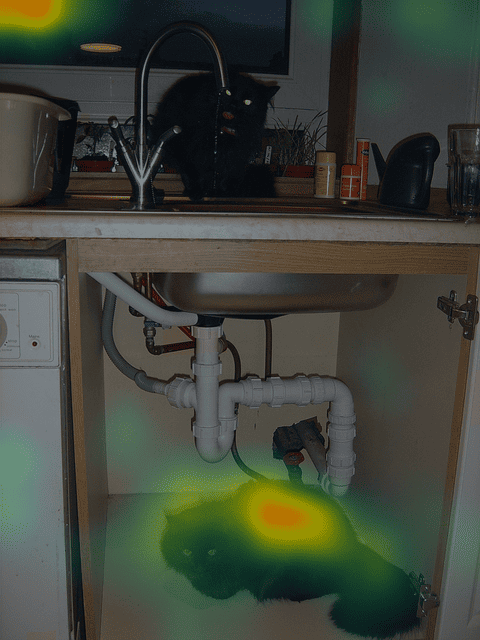} & 
\tokboxsmall{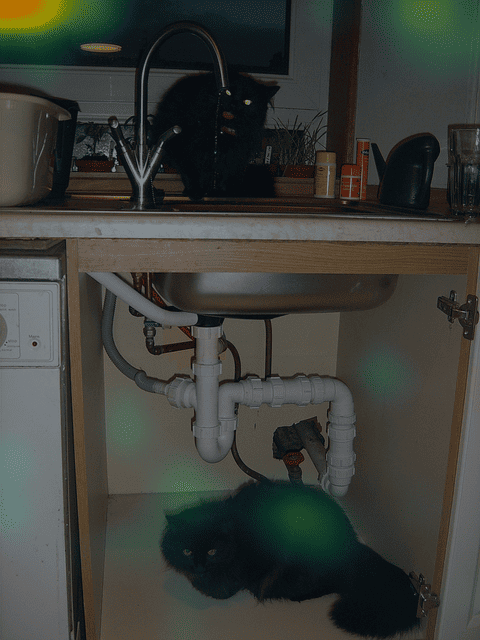} & 
\tokboxsmall{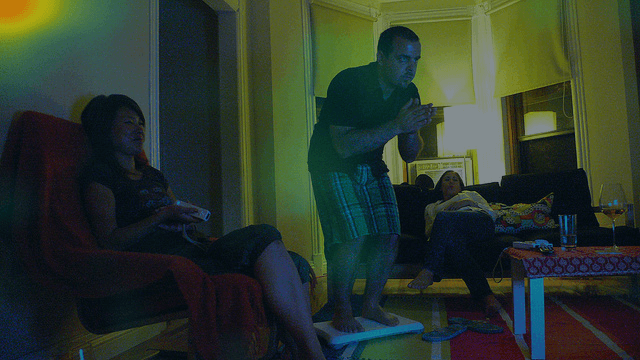} & 
\tokboxsmall{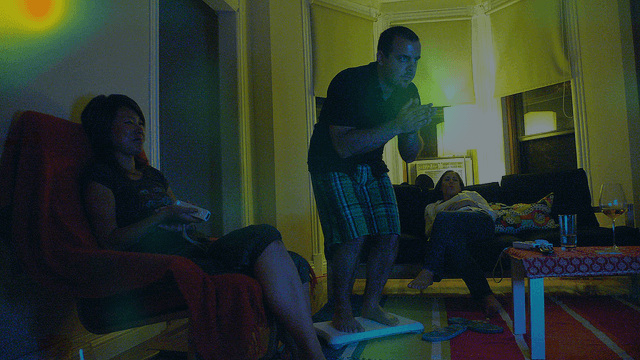} & 
\tokboxsmall{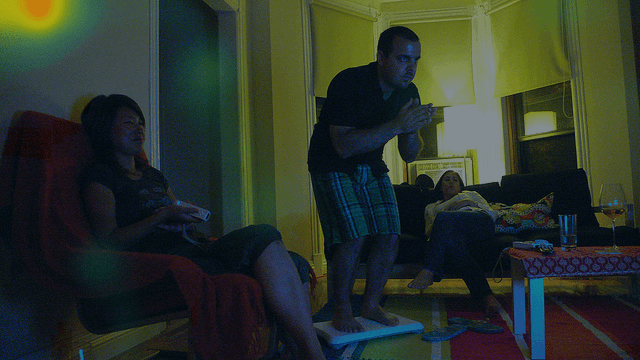} & 
\tokboxsmall{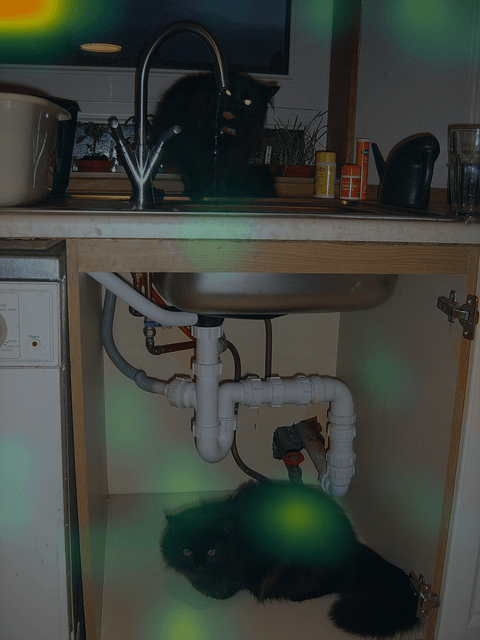} & 
\tokboxsmall{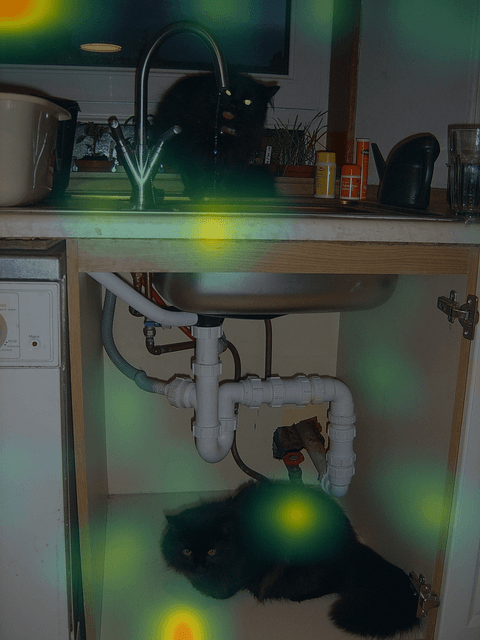} & 
\tokboxsmall{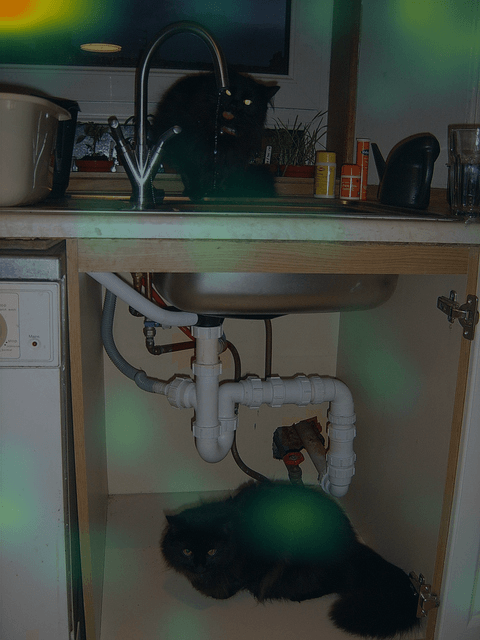} \\

\rotatebox{90}{{Semantic}} & 
\tokboxsmall{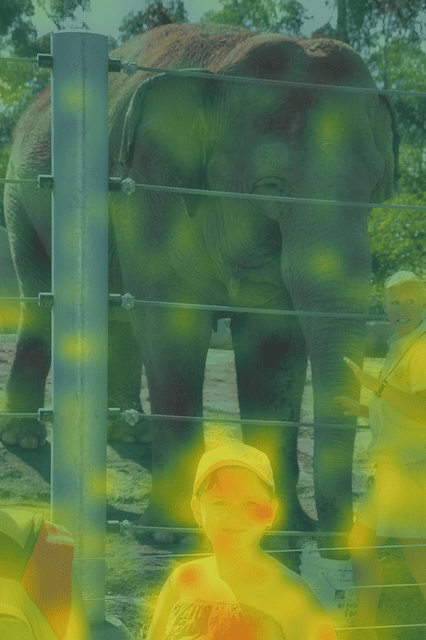} & 
\tokboxsmall{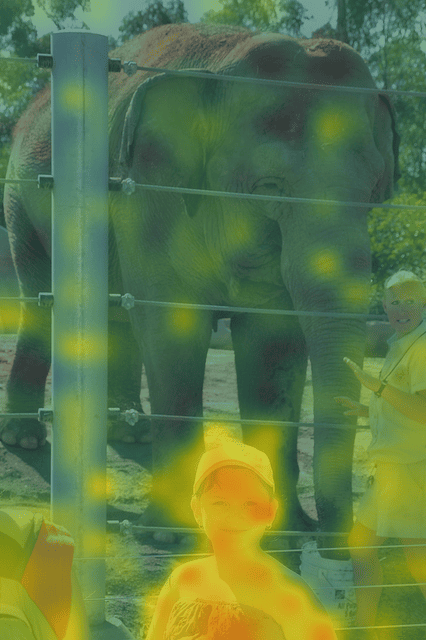} & 
\tokboxsmall{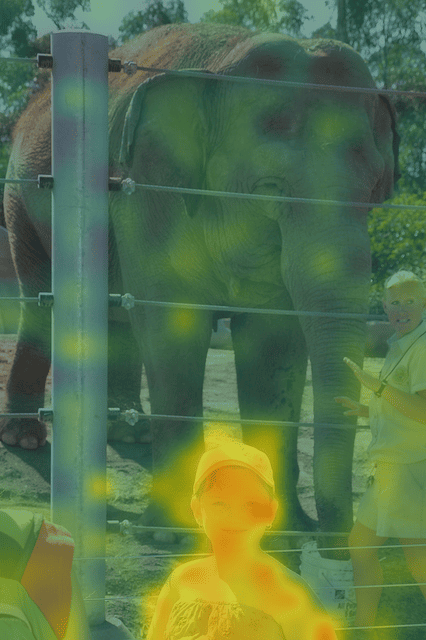} & 
\tokboxsmall{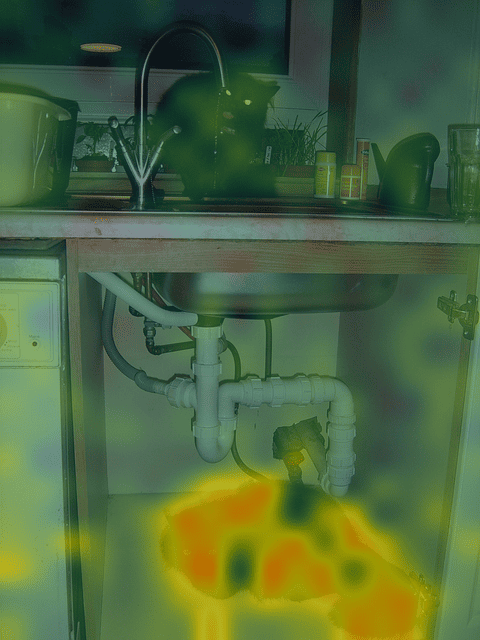} & 
\tokboxsmall{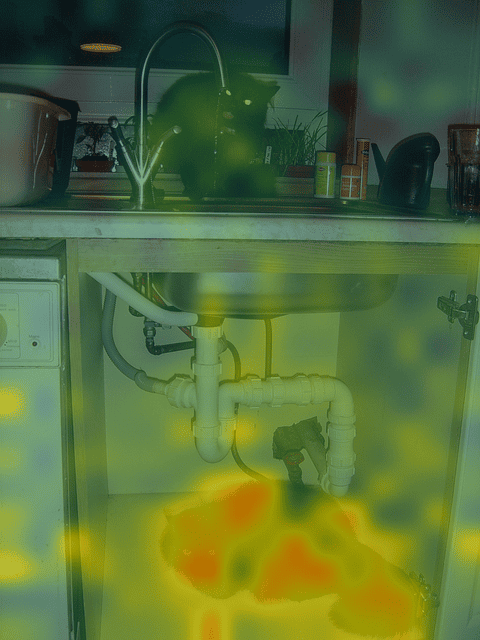} & 
\tokboxsmall{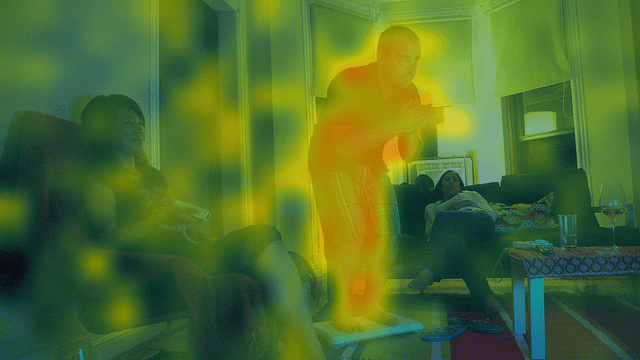} & 
\tokboxsmall{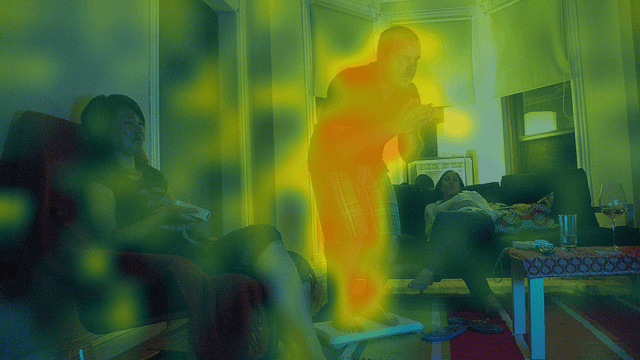} & 
\tokboxsmall{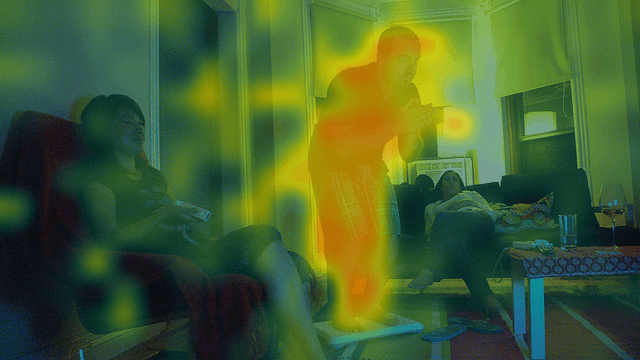} & 
\tokboxsmall{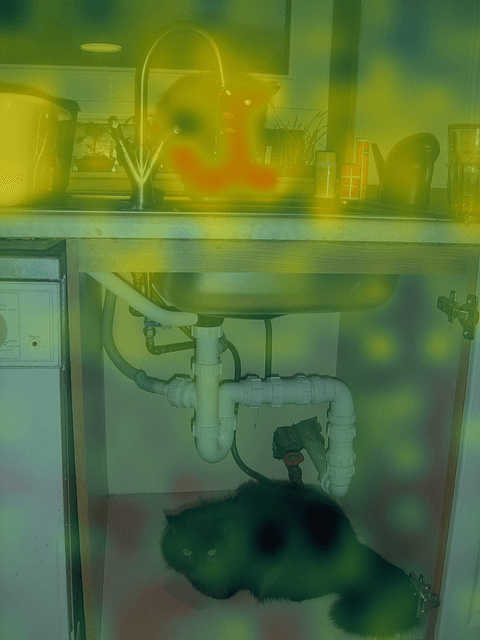} & 
\tokboxsmall{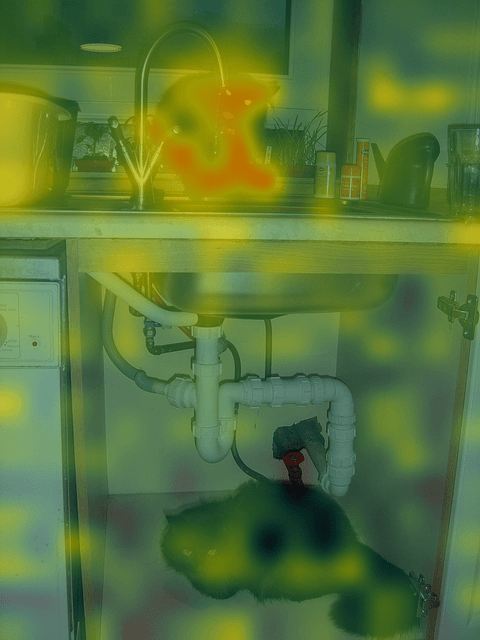} & 
\tokboxsmall{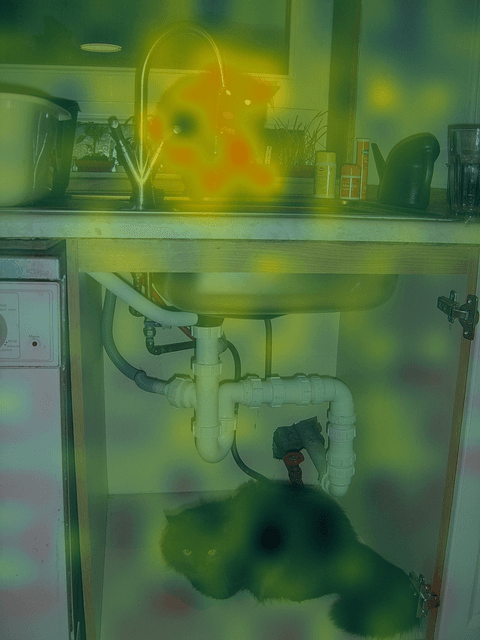} \\

\rotatebox{90}{{Center}} & 
\tokboxsmall{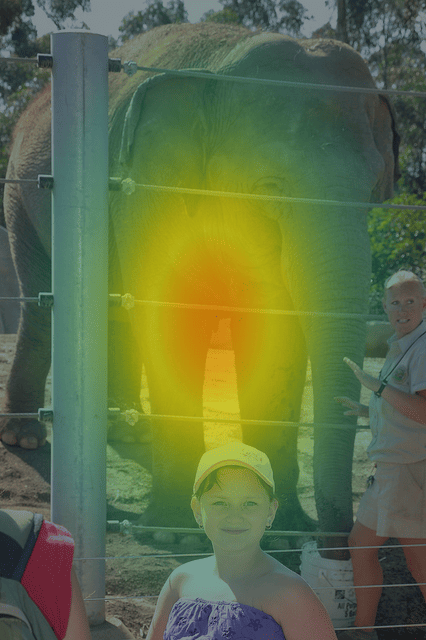} & 
\tokboxsmall{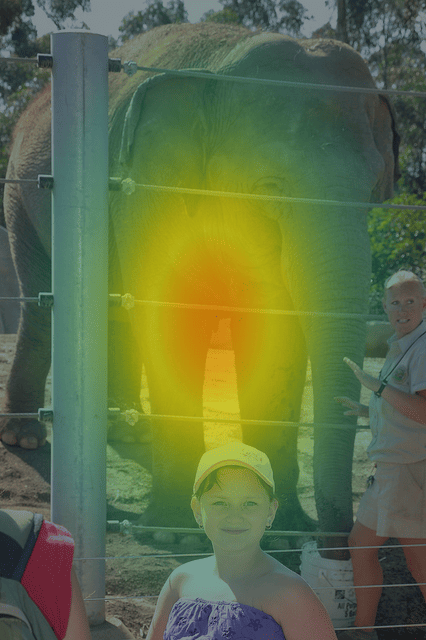} & 
\tokboxsmall{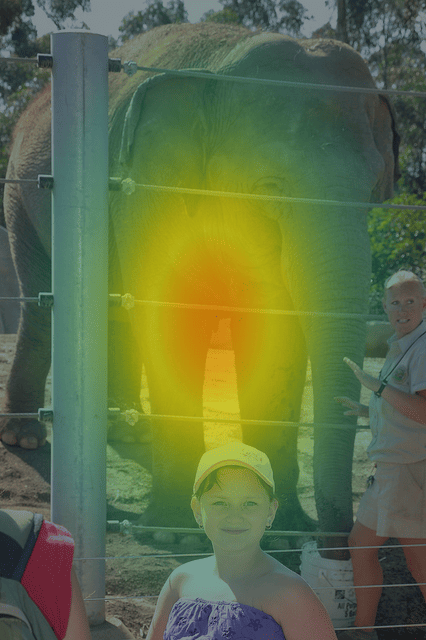} & 
\tokboxsmall{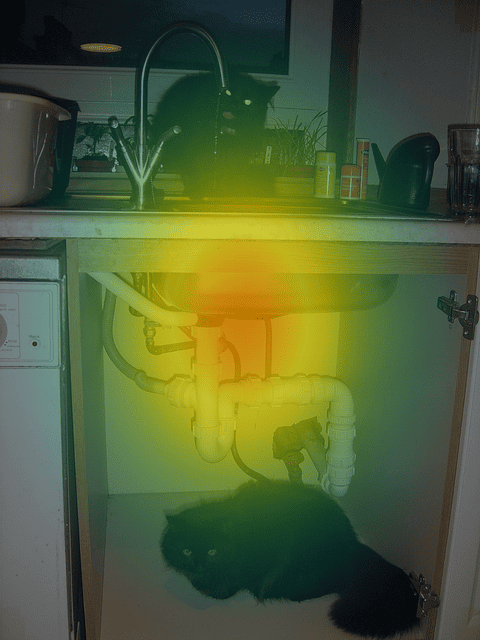} & 
\tokboxsmall{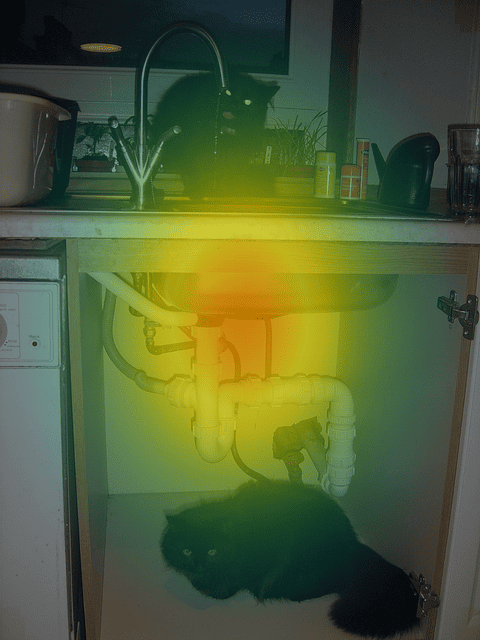} & 
\tokboxsmall{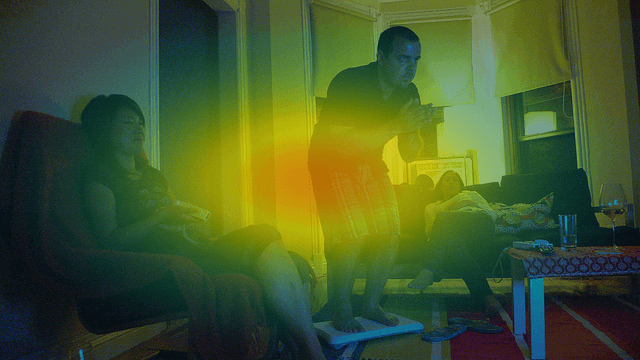} & 
\tokboxsmall{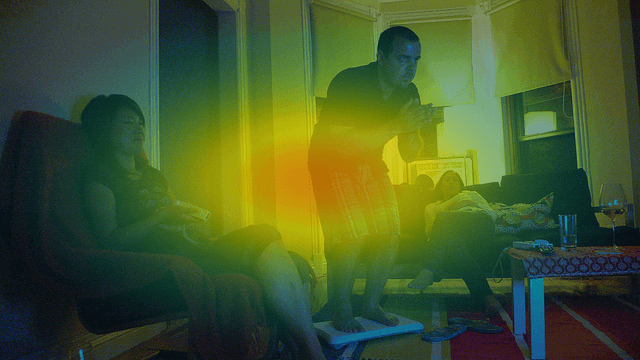} & 
\tokboxsmall{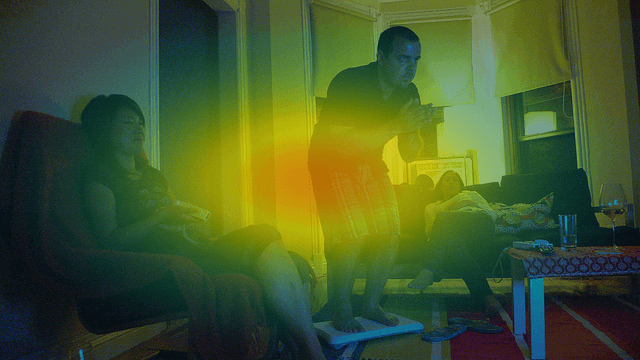} & 
\tokboxsmall{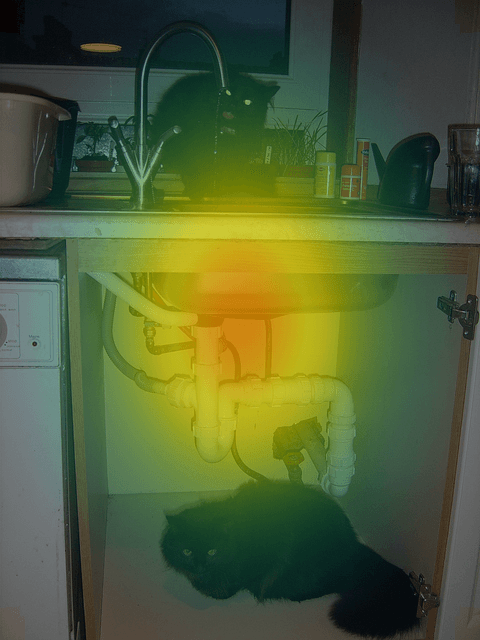} & 
\tokboxsmall{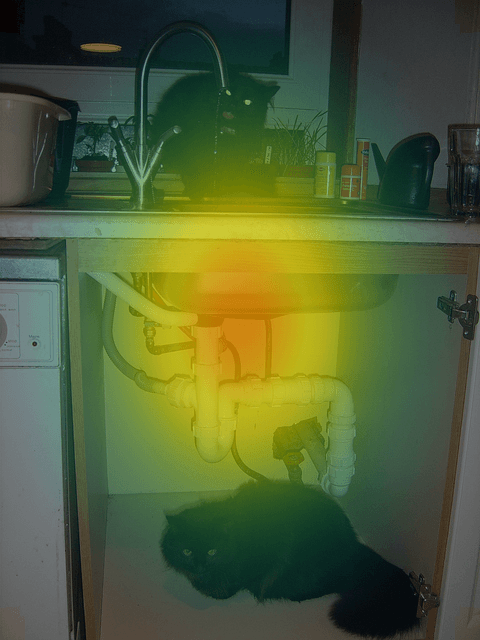} & 
\tokboxsmall{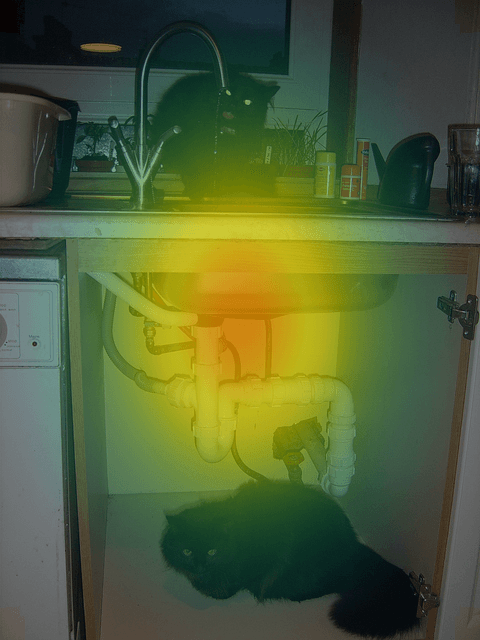} \\

\rotatebox{90}{{Random}} & 
\tokboxsmall{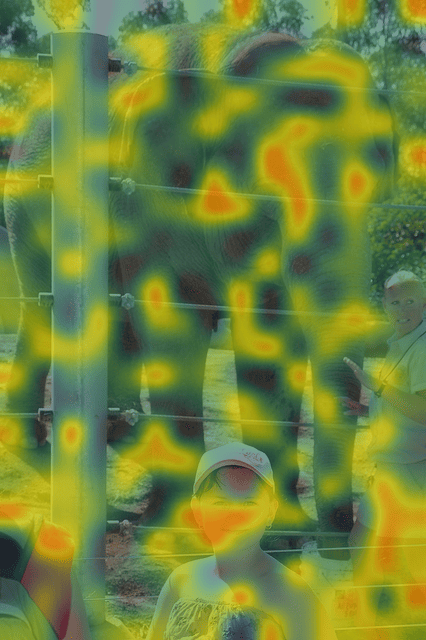} & 
\tokboxsmall{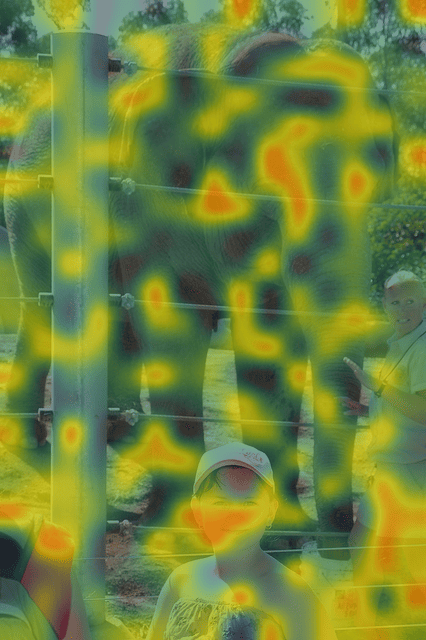} & 
\tokboxsmall{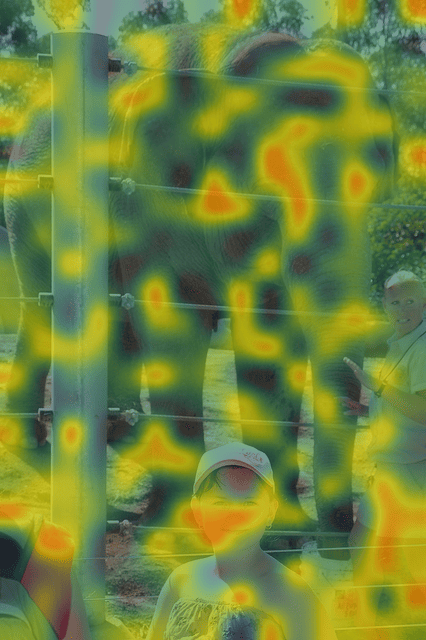} & 
\tokboxsmall{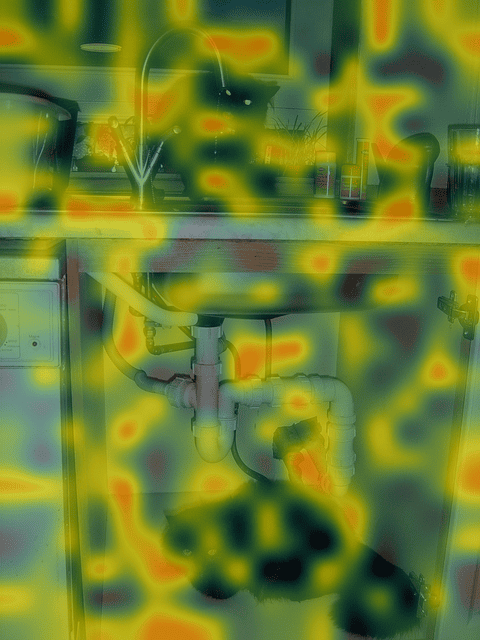} & 
\tokboxsmall{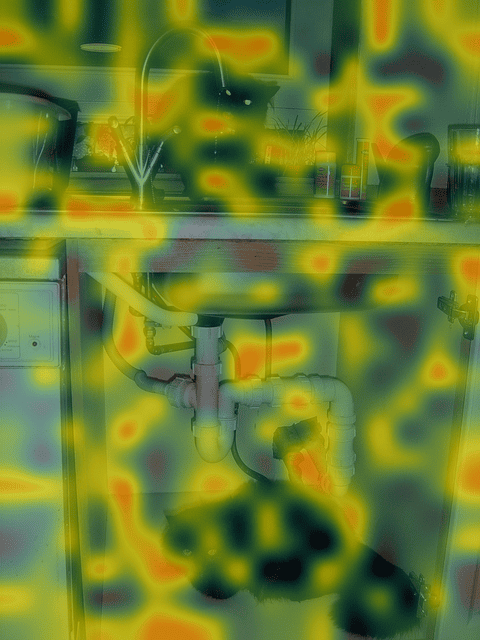} & 
\tokboxsmall{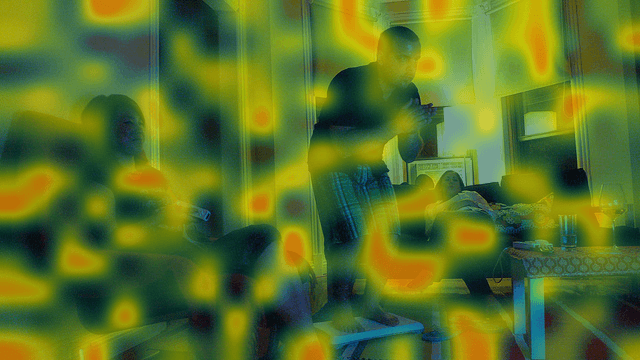} & 
\tokboxsmall{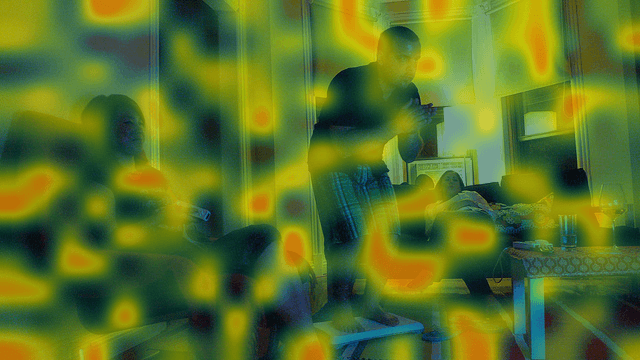} & 
\tokboxsmall{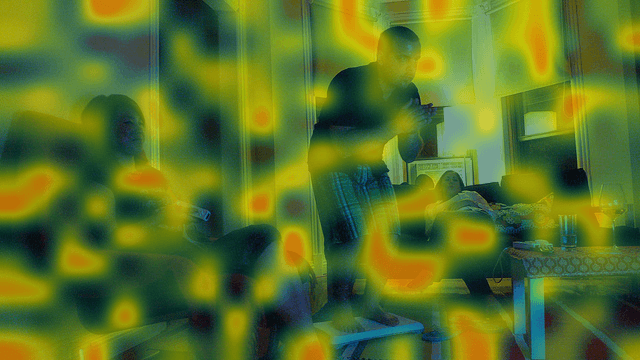} & 
\tokboxsmall{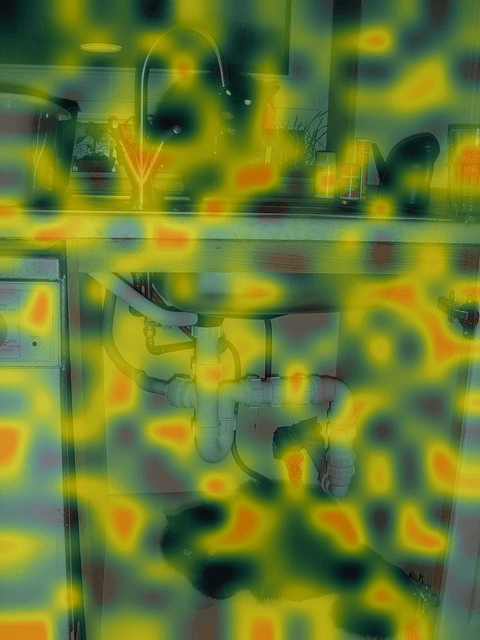} & 
\tokboxsmall{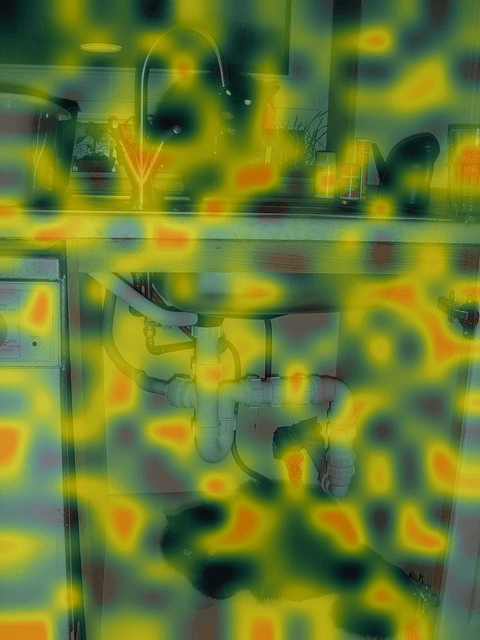} & 
\tokboxsmall{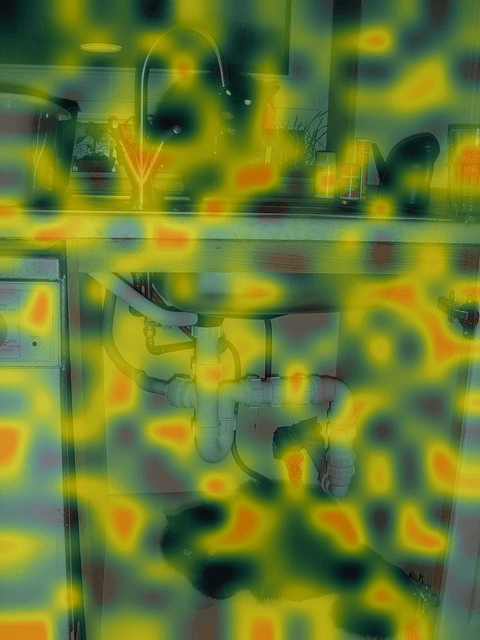} \\
\end{tabular}

\vspace{-2mm}
\caption{
Qualitative comparison on token-level grounding. 
We compare OTTER against TAM, attention, semantic similarity, center, and random baselines. 
}
\vspace{-4mm}
\label{fig:qualitative_token_grounding}
\end{figure*}
\section{Conclusion}
\label{sec:conclusion}

We studied the \emph{what-to-which} problem in frozen MLLMs: whether their
representations encode not only what visual entity a generated expression
describes, but also which instance of that entity is intended. Using controlled
referring expressions, counterfactual same-category pairs, context
perturbations, and free-generation settings, we find that instance-level
grounding is not confined to a single modifier token. Instead, it is distributed
across contextualized token representations: object nouns already carry
substantial information about the intended instance, while spatial modifiers
typically provide the strongest additional discrimination. This information
depends on both the visual input and the referring context, and remains
detectable when the expression is generated before the disambiguating modifier
is produced.
To analyze this phenomenon, we introduced OTTER, a lightweight supervised probe
that learns a token--region geometry over frozen MLLM representations. The
matched controls and perturbation experiments show that the recovered maps
capture more than raw feature similarity and identify regions relevant to the
model's generation. Moreover, the learned geometry remains useful under
unconstrained captioning and transfers to RefCOCOg without retraining. These
results suggest that the transition from \emph{what} to \emph{which} is better
understood as a process of progressively refined, distributed instance
grounding across an expression, rather than as information introduced only by
an isolated modifier. While our study covers a limited set of backbones,
datasets, and grounding settings, it provides a framework and empirical basis
for investigating this phenomenon across broader architectures and domains.
\clearpage

\bibliography{iclr2027_conference}
\bibliographystyle{iclr2027_conference}

\clearpage
\appendix

\section{Additional Architecture Ablation Details}
\label{app:architecture_ablation}

\subsection{Layer and Training-Scale Ablations}
\label{sec:appendix-layer-training-ablations}

\paragraph{Motivation.}
OTTER reads localization evidence from hidden states of a frozen MLLM. Since these representations are not equally spatial at all depths, we first screen the feature layer used by the segmentation head before comparing architectural variants. This control is especially important in the cross-backbone setting: a poor layer choice can make a backbone appear weak even when its intermediate states contain usable grounding information. We therefore separate layer selection from the final configuration ablation.

\paragraph{Cross-backbone layer screening.}
For each backbone, we freeze the MLLM and train only the OTTER head with the minimal \texttt{ot\_semantic\_only} configuration. We use RefCOCO, the training split for both head training and calibration with disjoint image ranges, and the validation split for evaluation. The layer-screening protocol uses 500 training images, 300 calibration images, and 1000 validation images; it keeps the first expression per mask, trains for up to 5 epochs with early stopping, and evaluates calibrated mIoU on the validation subset. Candidate layers are chosen at approximately 25\%, 50\%, 75\%, and 90\% of the language depth, plus the last two layers. Feature extraction uses sharded caches of 64 examples.

\begin{table}[ht]
\centering
\small
\setlength{\tabcolsep}{4pt}
\caption{Cross-backbone layer screening. Metrics are percentages. The selection score is the area-matched diagnostic IoU; calibrated mIoU is reported as the task metric. For each backbone, the selected layer is highlighted in \textbf{bold}.}
\label{tab:cross-backbone-layer-screen}
\begin{tabular}{lrrr}
\toprule
Backbone (layer) & Diag. IoU & mIoU & P@0.5 \\
\midrule
Qwen2.5-VL-3B L9 & 28.75 & 27.51 & 14.14 \\
Qwen2.5-VL-3B L18 & 38.45 & 35.23 & 29.69 \\
\textbf{Qwen2.5-VL-3B L27} & \textbf{44.22} & \textbf{41.06} & \textbf{40.92} \\
Qwen2.5-VL-3B L32 & 41.39 & 38.20 & 34.67 \\
Qwen2.5-VL-3B L35 & 28.76 & 28.56 & 11.88 \\
Qwen2.5-VL-3B L36 & 27.30 & 27.10 & 9.65 \\
\midrule
Qwen3-VL-2B L7 & 30.06 & 29.55 & 14.73 \\
\textbf{Qwen3-VL-2B L14} & \textbf{46.34} & \textbf{42.41} & \textbf{41.73} \\
Qwen3-VL-2B L21 & 42.20 & 37.40 & 31.87 \\
Qwen3-VL-2B L25 & 35.33 & 32.85 & 20.11 \\
Qwen3-VL-2B L27 & 26.82 & 26.57 & 8.08 \\
Qwen3-VL-2B L28 & 26.35 & 26.10 & 7.56 \\
\midrule
InternVL3.5-2B L7 & 10.07 & 13.93 & 1.11 \\
\textbf{InternVL3.5-2B L14} & \textbf{11.51} & \textbf{15.63} & \textbf{1.62} \\
InternVL3.5-2B L21 & 10.62 & 14.71 & 1.31 \\
InternVL3.5-2B L25 & 9.97 & 14.04 & 1.03 \\
InternVL3.5-2B L27 & 9.50 & 13.60 & 1.03 \\
InternVL3.5-2B L28 & 9.70 & 13.46 & 0.91 \\
\midrule
\textbf{Gemma-3-4B L8} & \textbf{10.44} & \textbf{13.96} & \textbf{0.71} \\
Gemma-3-4B L17 & 10.33 & 13.65 & 0.83 \\
Gemma-3-4B L26 & 9.87 & 13.26 & 0.67 \\
Gemma-3-4B L31 & 9.90 & 13.31 & 0.99 \\
Gemma-3-4B L33 & 9.90 & 13.71 & 1.03 \\
Gemma-3-4B L34 & 10.10 & 13.93 & 1.19 \\
\bottomrule
\end{tabular}
\end{table}

\paragraph{Mini token-type protocol for non-Qwen2.5 backbones.}
To avoid a full paper-scale rerun for every backbone, we use a small token-type check on RefCOCO modifier expressions. The test uses the best available head from the layer sweep for each backbone and evaluates \texttt{refcoco\_mod\_token\_groups} only. This is intended as a fast sanity check of token-level grounding (attributes, nouns, spatial terms, and relational/action terms), not as a final cross-backbone benchmark.

\begin{table}[ht]
\centering
\small
\setlength{\tabcolsep}{4pt}
\caption{Mini token-type grounding check on RefCOCO modifier expressions using sweep heads. 
Rows report per-token maps grouped by token type.}
\label{tab:mini-token-type-cross-backbone}
\begin{tabular}{lrrrrr}
\toprule
Backbone (layer) & Noun & Attribute & Spatial & Rel./act. & Token avg. \\
\midrule
Qwen2.5-VL-3B L27 & 37.91 & 39.47 & 39.05 & 32.75 & 37.30 \\
Qwen3-VL-2B L14 & 29.95 & 36.82 & 31.51 & 32.34 & 32.66 \\
InternVL3.5-2B L14 & 12.19 & 12.29 & 12.82 & 10.23 & 11.88 \\
Gemma-3-4B L17 & 12.08 & 10.93 & 10.48 & 9.45 & 10.74 \\
\bottomrule
\end{tabular}
\end{table}

\paragraph{Training-scale and configuration ablation.}
After layer screening, we rerun the main Qwen2.5-VL-3B experiment at the selected layer. This experiment uses all RefCOCO validation expressions (10,707 expression--mask pairs), 300 calibration images, and training subsets of increasing size. Unlike the layer screen, this run uses \texttt{example\_mode=all\_pairs}, so each referring expression is an independent training/evaluation example. The OT head is trained for up to 20 epochs with early stopping on calibration mIoU, and thresholds are calibrated on \{0.1, 0.2, 0.3, 0.4, 0.5\}. We compare softmax normalization against OT variants with the same semantic, attention, and spatial cost terms.

\begin{table}[ht]
\centering
\scriptsize
\setlength{\tabcolsep}{3.0pt}
\renewcommand{\arraystretch}{1.08}
\caption{Combined architecture and training-scale ablation at Qwen2.5-VL-3B layer 27 on RefCOCO validation. The left side reports the mIoU across increasing numbers of requested training images. The right side reports detailed metrics (mIoU, cIoU, P@0.5, Diag. IoU) for all configurations at the largest training scale (2500 images). The overall best configuration is highlighted in \textbf{bold}.}
\label{tab:ntrain-scaling-layer27}
\resizebox{\linewidth}{!}{%
\begin{tabular}{lrrrrrrr|rrrr}
\toprule
& \multicolumn{7}{c|}{mIoU at Training Scale (Images)} & \multicolumn{4}{c}{Full Metrics at 2500 Images} \\
Variant & 50 & 100 & 300 & 600 & 900 & 1500 & 2000 & mIoU & cIoU & P@0.5 & Diag. IoU \\
\midrule
\texttt{softmax\_semantic\_only} & 21.83 & 23.81 & 31.31 & 37.51 & 40.36 & 42.09 & 42.96 & 43.74 & 42.81 & 45.79 & 48.65 \\
\texttt{softmax\_no\_ot} & 22.08 & 24.76 & 29.42 & 38.17 & 40.36 & 41.87 & 42.17 & 43.74 & 43.19 & 45.52 & 49.15 \\
\texttt{softmax\_fixed\_full} & 22.24 & 24.83 & 29.04 & 38.76 & 40.44 & 42.36 & 41.84 & 43.69 & 42.58 & 45.42 & 50.48 \\
\midrule
\texttt{ot\_semantic\_only} & 21.02 & 24.53 & 34.47 & 39.69 & 40.50 & 43.26 & 44.74 & 45.66 & 46.08 & 49.74 & 49.57 \\
\texttt{ot\_sem\_spatial} & 20.84 & 24.11 & 34.12 & 39.63 & 41.70 & 42.92 & 44.63 & 45.12 & 45.41 & 47.97 & 49.67 \\
\texttt{ot\_sem\_att} & 21.81 & 25.00 & 33.94 & 39.40 & 41.41 & 43.29 & 44.41 & 45.17 & 45.43 & 49.41 & 51.20 \\
\midrule
\texttt{ot\_full} & 22.28 & 25.59 & 33.97 & 39.41 & 41.69 & 43.10 & 44.32 & 44.94 & 45.25 & 49.11 & 51.26 \\
\texttt{ot\_full\_fixed\_dustbins} & 22.59 & \textbf{26.45} & 34.46 & 39.59 & 41.78 & 43.10 & 44.30 & 44.99 & 45.31 & 49.29 & 51.09 \\
\texttt{ot\_full\_learned\_weights} & \textbf{22.85} & 26.02 & 33.92 & 39.52 & 41.79 & \textbf{43.60} & \textbf{44.85} & 45.44 & 45.55 & 49.14 & 51.30 \\
\midrule
\texttt{ot\_adaptive\_gated} & 22.34 & 25.79 & \textbf{34.63} & \textbf{40.41} & \textbf{41.88} & 43.12 & 44.71 & 45.61 & 45.60 & 50.17 & 51.21 \\
\textbf{\texttt{ot\_adaptive\_gated\_learned\_weights}} & 22.27 & 24.99 & 33.95 & 39.80 & 41.68 & 43.21 & 44.73 & \textbf{45.83} & \textbf{46.05} & \textbf{50.66} & \textbf{51.32} \\
\bottomrule
\end{tabular}}
\end{table}

\paragraph{Results.}
The layer screen shows that localization is strongly layer-dependent. For Qwen2.5-VL-3B, layer 27 (75\% depth) is best among the screened layers, reaching 41.06 mIoU and 44.22 diagnostic IoU on the 1000-image screen. Qwen3-VL-2B peaks earlier, at layer 14 (50\% depth), with 42.41 mIoU and 46.34 diagnostic IoU. Gemma-3-4B remains weak across screened depths: its best official calibrated mIoU is at layer 8. 

The mini token-type protocol confirms that Qwen3-VL-2B transfers reasonably well to token-level modifier grounding: with the layer-14 semantic-only head it reaches a 32.66 average token-type soft-IoU, with especially strong attribute grounding (36.82). In contrast, InternVL3.5-2B and Gemma-3-4B remain close to 12 and 11 average token-type soft-IoU, respectively, under the same conditions, suggesting that the current generic adapter/head combination does not yet expose comparably useful token-to-patch geometry for these backbones.

At the selected Qwen2.5-VL layer, OT-based heads consistently outperform matched softmax baselines at moderate and large training scales. At 600 training images, the best softmax variant reaches 38.76 mIoU, whereas adaptive-gated OT reaches 40.41 mIoU. The advantage grows with more supervision: at the largest requested scale, the best softmax heads reach 43.74 mIoU, while adaptive-gated OT with learned weights reaches 45.83 mIoU. Semantic-only OT is already a strong low-complexity reference, reaching 45.66 mIoU at the largest scale, but the adaptive-gated learned-weight variant gives the best final calibrated mIoU. These results support two design choices used in the main system: selecting an intermediate representation layer by validation grounding diagnostics, and using OT rather than independent softmax normalization for map construction.

\subsection{Why Optimal Transport? A Controlled Operator Ablation}
\label{sec:why_ot}

The architecture ablation shows OT heads above softmax heads on segmentation mIoU, but mIoU alone does not explain \emph{why} transport helps. We isolate the transport operator by holding the learned geometry and the cost terms fixed and switching the aggregation---Sinkhorn optimal transport with dustbins versus a softmax pooling---and we evaluate on the dense grounding metrics that matter for a token-level map: the target-vs-distractor gap, Soft-IoU, and map entropy, computed per token group on RefCOCO validation.

Two comparisons are controlled: at the semantic-only cost (\texttt{ot\_semantic\_only} vs.\ \texttt{softmax\_semantic\_only}) and at the full cost (\texttt{ot\_full} vs.\ \texttt{softmax\_no\_ot}). Both use the same cost terms and the same initial weights ($w_{\mathrm{sem}}{=}1.0$, $w_{\mathrm{att}}{=}0.5$, $w_{\mathrm{spa}}{=}0.2$). Two secondary differences accompany the operator and both cut in OT's favor: the softmax head additionally \emph{learns} its cost weights, whereas the OT head keeps them fixed, and dustbin abstention is available only under transport because it is structurally incompatible with a normalized softmax. Thus the softmax head is, if anything, the more flexible of the two on cost weighting.

\begin{table}[ht]
\centering
\scriptsize
\caption{Controlled operator ablation with $95\%$ paired bootstrap CIs ($n$ per group). Concentration is reported as peak-to-mean ratio; higher is more concentrated. This non-saturating metric reveals the operator effect that normalized entropy hides. \emph{Effect of OT at full cost} isolates the Sinkhorn operator; \emph{cost terms exploited} shows the gain from adding attention+spatial to each operator. All intervals shown exclude zero. Gap is undefined for nouns because there are no same-category instance distractors.}
\label{tab:why_ot}
\begin{tabular}{llccc}
\toprule
& Token group & $\Delta$Soft-IoU & $\Delta$gap & $\Delta$peak/mean \\
\midrule
\multirow{3}{*}{\shortstack[l]{Effect of OT\\at full cost\\{\footnotesize(\texttt{ot\_full}$-$\texttt{sm\_no\_ot})}}}
& Spatial   & $+0.021\,[.019,.023]$ & $-0.018\,[\text{-}.022,\text{-}.013]$ & $\mathbf{+5.41\,[5.21,5.62]}$ \\
& Attribute & $+0.018\,[.015,.021]$ & $-0.028\,[\text{-}.033,\text{-}.022]$ & $\mathbf{+5.93\,[5.65,6.22]}$ \\
& Noun      & $+0.026\,[.025,.027]$ & --- & $\mathbf{+3.85\,[3.75,3.96]}$ \\
\midrule
\multicolumn{5}{l}{\emph{Cost terms exploited: gain from adding attention+spatial to the semantic cost}}\\
\multirow{2}{*}{\shortstack[l]{OT uses them\\{\footnotesize(\texttt{ot\_full}$-$\texttt{ot\_sem})}}}
& Spatial   & $+0.024\,[.022,.026]$ & $+0.031\,[.027,.036]$ & $+0.77\,[.64,.90]$ \\
& Attribute & $+0.022\,[.019,.026]$ & $+0.030\,[.023,.036]$ & $-1.32\,[\text{-}1.59,\text{-}1.06]$ \\
\multirow{2}{*}{\shortstack[l]{Softmax wastes them\\{\footnotesize(\texttt{sm\_no\_ot}$-$\texttt{sm\_sem})}}}
& Spatial   & $+0.001\,[.000,.001]$ & $+0.004\,[.004,.005]$ & $-0.06\,[\text{-}.08,\text{-}.04]$ \\
& Attribute & $+0.002\,[.002,.003]$ & $+0.006\,[.005,.007]$ & $-0.22\,[\text{-}.24,\text{-}.20]$ \\
\bottomrule
\end{tabular}
\end{table}

\paragraph{Optimal transport makes multi-term costs usable.}
The most decisive result is in the lower half of Table~\ref{tab:why_ot}. Adding attention and spatial cost terms on top of the semantic cost improves the maps substantially \emph{under OT}: gap $+0.031$ for spatial, $+0.030$ for attributes, and Soft-IoU about $+0.02$. Under softmax, the same additions barely move the metrics: gap $+0.004$ for spatial and $+0.006$ for attributes. 

The paired bootstrap intervals do not overlap: OT exploits the extra cost terms roughly $5$--$7\times$ as much as softmax on the gap, significant on every group and metric. This holds even though the softmax head is allowed to \emph{learn} its cost weights while the OT head keeps them fixed. Given the freedom to reweight attention and spatial terms, softmax still does not use them, whereas fixed-weight OT does. The advantage is therefore not a weighting artifact; it is the assignment structure of transport. This is a structural reason to prefer OT: it integrates competing semantic, attention, and spatial signals through a global assignment, whereas independent softmax normalization dilutes them. It also explains why \texttt{softmax\_no\_ot} and \texttt{softmax\_semantic\_only} score almost identically despite different cost terms: the extra terms are present but unused.

\paragraph{At full cost, OT is more accurate and more compact.}
Isolating the operator at full cost, the top block of Table~\ref{tab:why_ot} shows that OT yields significantly higher Soft-IoU, about $+0.02$, and much more concentrated maps. The peak-to-mean ratio rises by $+3.9$ to $+5.9$, a $50$--$90\%$ relative increase over softmax, with tight non-overlapping intervals. This concentration advantage is a property of the transport operator itself, not of the extra cost terms: it is large in the operator-only comparisons, but small or absent in the cost-term comparisons, where attention and spatial mainly improve gap and Soft-IoU. 

We report peak-to-mean ratio because normalized entropy understates the same effect: it moves only about $-0.02$ on identical maps, saturating in this high-entropy regime. The one axis where softmax leads is the raw target-vs-distractor gap, with OT lower by $0.018$ on spatial and $0.028$ on attributes. Softmax places a slightly sharper single peak on the target but distributes the rest more diffusely, which is why its region overlap and concentration are worse. The complementary structural argument is the dustbin: OT can route non-groundable tokens to a sink and abstain, whereas softmax must place all mass on the image. We note that the full-cost OT head carries learned dustbins while the softmax head has none, so the accuracy and compactness advantages above bundle transport assignment with dustbin abstention rather than isolating transport alone. Since abstention is itself only available under transport, we treat the two as jointly constitutive of the operator rather than as separable confounds.

\end{document}